\PassOptionsToPackage{table}{xcolor}
\documentclass[11pt]{article}

\usepackage[final]{acl}

\usepackage{times}
\usepackage{latexsym}
\usepackage[T1]{fontenc}
\usepackage[utf8]{inputenc}
\usepackage{microtype}
\usepackage{inconsolata}
\usepackage{graphicx}

\usepackage{amsmath,amssymb,amsthm}

\usepackage{booktabs}
\usepackage{multirow}
\usepackage{makecell}
\usepackage{array}

\usepackage{algorithm}
\usepackage{algpseudocode}
\algnewcommand\algorithmicinput{\textbf{Input:}}
\algnewcommand\algorithmicoutput{\textbf{Output:}}
\algnewcommand\Input{\item[\algorithmicinput]}
\algnewcommand\Output{\item[\algorithmicoutput]}

\usepackage{enumitem}

\usepackage{xcolor}
\definecolor{myblue}{RGB}{0,80,160}
\definecolor{mygray}{RGB}{130,130,130}

\usepackage{siunitx}
\usepackage{tikz}
\usetikzlibrary{arrows.meta,positioning,calc,fit,backgrounds}
\usepackage{pgfplots}
\pgfplotsset{compat=1.18}

\newtheorem{definition}{Definition}

\newcommand{\hvis}{H_{\text{vis}}}
\newcommand{\htxt}{H_{\text{txt}}}
\newcommand{\vig}{\text{VIG}}
\newcommand{\rvig}{R_{\text{VIG}}}
\newcommand{\racc}{R_{\text{acc}}}
\newcommand{\rfmt}{R_{\text{format}}}

\title{VIG: Visual Information Gain as a Reward Signal\\
       for Multimodal Chain-of-Thought Compression}

\author{
  Wen Luo\textsuperscript{1}\thanks{\;Equal contribution.} \quad
  Xiaohan Yi\textsuperscript{2}\footnotemark[1] \quad
  Xiaotao Huang\textsuperscript{1}\thanks{\;Corresponding authors.} \quad
  Liqun Huang\textsuperscript{1}\footnotemark[2] \\
  \textsuperscript{1}School of Software Engineering,
  Huazhong University of Science and Technology \\
  \textsuperscript{2}Tsinghua University \\
  \texttt{\{wen\_chaser, huangliqun\}@hust.edu.cn} \quad
  \texttt{huangxt@mail.hust.edu.cn} \\
  \texttt{yxh24@mails.tsinghua.edu.cn}
}

\begin{document}
\maketitle

\begin{abstract}
Multimodal large reasoning models often rely on long Chain-of-Thought (CoT)
traces in which a substantial fraction of tokens, such as repeated visual
descriptions, self-reflection, and other visually-disengaged filler, inflate
inference cost without contributing to the answer.
Existing CoT compression methods optimize output length but never measure
whether a reasoning token is actually grounded in the image.
We propose \textbf{VIG} (Visual Information Gain), an information-theoretic
GRPO reward that scores each reasoning token by how much the image reduces
its predictive uncertainty.
VIG is computed online from two forward passes of the same policy, one with
and one without the image, so no reference chains, external annotations, or
auxiliary reward models are needed.
Across six main multimodal reasoning benchmarks and three Qwen3-VL-Thinking
model sizes (2B/4B/8B), plus an additional R1-Onevision-Bench evaluation on
8B, VIG consistently improves the accuracy--efficiency trade-off, supporting
our central claim:
\emph{efficient multimodal reasoning emerges from raising visual information
density, where every reasoning token earns its place by anchoring to the
image, rather than from imposing a length budget.}
Our source code is available at
\url{https://github.com/chaser682/vig}.
\end{abstract}

\section{Introduction}
\label{sec:intro}

\begin{figure}[t]
\centering
\includegraphics[width=0.98\columnwidth]{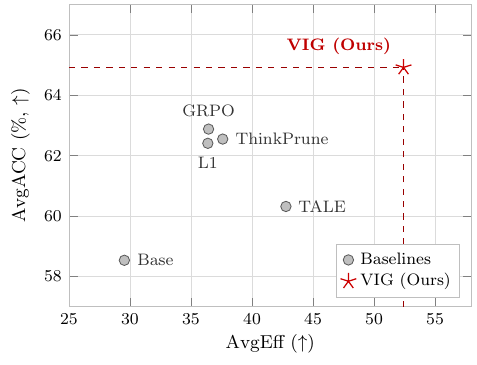}
\caption{\textbf{Accuracy--efficiency Pareto frontier on Qwen3-VL-8B-Thinking.}
Each point reports the average accuracy (AvgACC, \%) and efficiency
(AvgEff, ACC per 1k tokens) of one method across the six multimodal benchmarks
in Table~\ref{tab:main}.
VIG ({\color{red!75!black}$\bigstar$}) Pareto-dominates the evaluated baselines
(Base, GRPO, L1, ThinkPrune, TALE) on both axes, defining a new
accuracy--efficiency Pareto frontier (red dashed lines).}
\label{fig:pareto}
\end{figure}

A modern Multimodal Large Reasoning Model (MLRM) may emit over a thousand tokens of
``thinking'' to answer a single geometry question---yet a substantial fraction of those
tokens never actually consult the image.
They restate the question, narrate the solver's own metacognition, and rephrase visual
relationships already established earlier in the chain.
The result is a reasoning trace that is both wasteful to compute and, as we will show,
often \emph{less} accurate than a shorter, more visually grounded reasoning chain.

This phenomenon is a direct consequence of how deep reasoning is currently trained.
Inspired by OpenAI o1~\citep{openai_o1_2024} and
DeepSeek-R1~\citep{deepseekr1_2025}, modern
MLRMs~\citep{qwenvl2023,llava2023,internvl2024,bai2025qwen3} produce long
Chain-of-Thought (CoT) within \texttt{<think>} blocks, articulating visual details,
spatial relationships, and cross-modal alignments before committing to an answer.
Longer chains often correlate with higher accuracy on text-only benchmarks, but in
multimodal scenarios this paradigm comes with two costs.
First, the inflated token count drives up inference latency, restricting deployment in
interactive settings.
More surprisingly---and more importantly---over-reasoning actively harms correctness:
redundant steps inject misleading context, small perceptual errors accumulate across
hundreds of tokens, and models that initially read the image correctly are often
talked out of correct conclusions by their own elaboration.

\begin{figure*}[t]
\centering
\includegraphics[width=0.97\textwidth]{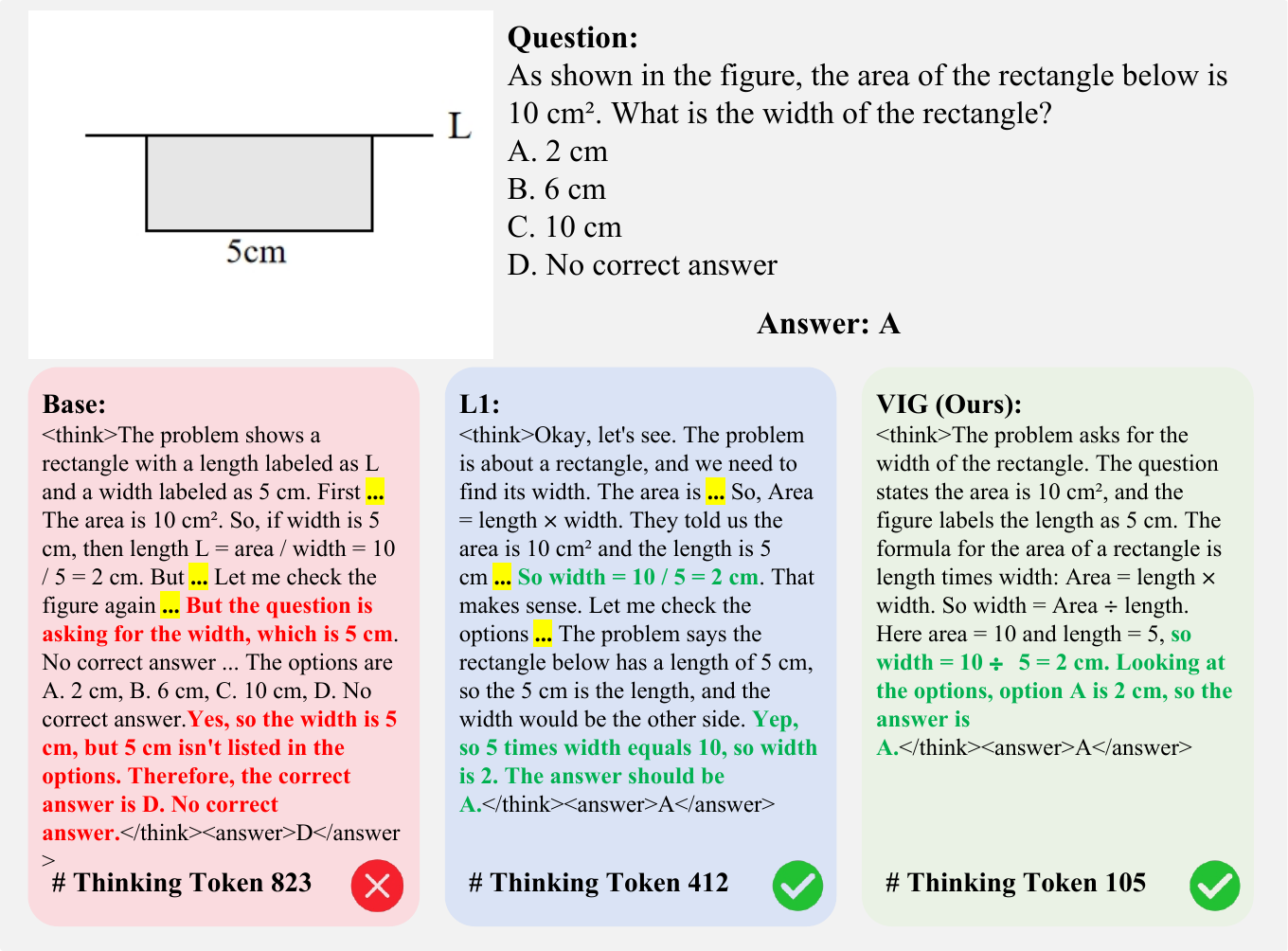}
\caption{\textbf{Length penalties trim the chain; only visually grounded
  reasoning removes the redundancy at its source.}
  Token counts and final answers from Base, L1, and our VIG model on the same
  geometry question (gold answer A, $2\,\text{cm}$).}
\label{fig:intro}
\end{figure*}

Figure~\ref{fig:intro} makes this concrete on an elementary geometry problem
(rectangle of area $10\,\text{cm}^2$ with one side labeled $5\,\text{cm}$):
Qwen3-VL-8B-Thinking~\citep{bai2025qwen3} spends 823 tokens and still answers
``no correct answer'', because its long deliberation second-guesses the
figure's plain label and talks itself out of the obvious division.
L1's length penalty~\citep{l1_2025} cuts the chain to 412 tokens and
recovers the correct answer, but still uses nearly $4\times$ the tokens of
our method---it truncates self-doubt without teaching the model to trust the
image.
Our VIG model finishes in 105 tokens by simply reading $5\,\text{cm}$ from
the figure and applying $\text{area}=\text{length}\times\text{width}$.
This is the blind spot shared by current CoT compression methods---prompting
~\citep{han2025token,kimi_k1_5_2025}, distillation on compressed
traces~\citep{li2026making,tokenskip_2025,cotvalve_2025}, and RL with
length, token-limit, or preference penalties~\citep{l1_2025,hou2025thinkprune,lcpo_2025}---and
recent multimodal reasoners~\citep{mmeureka_2025,visionr1_2025,vppo_2025,r1onevision_2025}
inherit it: every compression signal in the literature is purely textual,
unable to tell a visually-essential observation from linguistic filler.
What is missing is a per-token signal that measures \emph{how much each
token actually depends on the image.}

For each reasoning token, we ask exactly that: how much does the image
reduce the model's uncertainty about it.
Tokens that require the image---a direct reading of a value, a spatial
relationship, an identification of a visual entity---receive high scores;
tokens derivable from language context alone---meta-cognitive openers,
problem restatements, generic algebra---receive scores near zero.
We call this quantity Visual Information Gain (VIG); it coincides
with the conditional mutual information between the token and the
image~\citep{cover2006elements}.
Computing it requires only two forward passes of the same policy---one with
the visual tokens, one without---whose per-token entropy difference,
averaged over the \texttt{<think>} block, becomes a scalar reward plugged
directly into GRPO~\citep{grpo2024}.
No external annotator, reference chain, step segmentation, or auxiliary
model is needed.
Trained with VIG, the model learns to make every token earn its place by
anchoring to the image, with chain shortening emerging as a byproduct of
optimizing visual information density rather than of a length constraint.

Empirically, this principle yields consistent gains.
Across six multimodal reasoning benchmarks and three Qwen3-VL-Thinking model
sizes, VIG sets a new accuracy--efficiency Pareto frontier.
As shown in Figure~\ref{fig:pareto}, on Qwen3-VL-8B-Thinking VIG
Pareto-dominates the evaluated baselines---Base, plain GRPO~\citep{grpo2024},
L1~\citep{l1_2025}, ThinkPrune~\citep{hou2025thinkprune}, and
TALE~\citep{han2025token}---in both
accuracy and efficiency, reaching 64.92\% average accuracy and an efficiency
(ACC per 1k tokens) of 52.42 and beating the strongest compression baseline
(ThinkPrune) by $+2.37\%$ in accuracy and $+39.4\%$ in efficiency.
The gains extend to R1-Onevision-Bench~\citep{r1onevision_2025}, where VIG
ranks first overall and first or tied-first on three of five subject
categories, confirming that the benefit of visual-density-aware optimization
is not specific to a particular benchmark family.

\paragraph{Contributions.}
\begin{enumerate}[nosep,leftmargin=1.2em]
  \item We introduce \textbf{Visual Information Gain}, a per-token,
        information-theoretic measure of how much an MLRM's reasoning token
        actually relies on the visual input, reframing CoT compression as
        increasing visual density rather than reducing length.
  \item We turn VIG into a fully online RL reward computable from two forward passes
        of the policy itself---no external models, no annotated traces, no step
        segmentation---dropping cleanly into GRPO with no additional supervision.
  \item On Qwen3-VL-8B-Thinking, VIG delivers $+39.4\%$ efficiency and $+2.37\%$
        accuracy over the strongest compression baseline; across all three
        Qwen3-VL-Thinking model sizes it lies on the main-benchmark Pareto
        frontier, and on R1-Onevision-Bench it ranks first overall and first or
        tied-first on three of five subject categories.
\end{enumerate}

\section{Related Work}
\label{sec:related}

\subsection{Multimodal Large Reasoning Models}
\label{ssec:related_mlrm}

RL-based deep reasoning in LLMs~\citep{deepseekr1_2025,openai_o1_2024} has
quickly been extended to multimodal settings.
Early systems such as LLaVA-CoT~\citep{llavacot_2024} and
Mulberry~\citep{mulberry_2024} introduced structured visual reasoning via
stage-wise decomposition and search.
Recent RL-based MLRMs further scale long-CoT reasoning in vision-language tasks:
VisualThinker-R1-Zero~\citep{visualthinkerr1zero_2025} and
MM-Eureka~\citep{mmeureka_2025} study R1-style reasoning and ``aha-moment''
behavior, while OThink-MR1~\citep{othinkmr1_2025},
ThinkLite-VL~\citep{thinklitevl_2025}, Vision-R1~\citep{visionr1_2025},
LMM-R1~\citep{lmmr1_2025}, R1-Omni~\citep{r1omni_2025},
Insight-V~\citep{insightv_2024}, and
R1-OneVision~\citep{r1onevision_2025} improve training recipes and broaden the
scope of multimodal reasoning.
However, stronger reasoning often comes with substantially longer CoTs, including
repeated visual descriptions and unnecessary self-reflection.
Closest to ours is VPPO~\citep{vppo_2025}, which reweights trajectory advantages
by overall visual dependency; in contrast, VIG targets explicit CoT compression
by turning a with-/without-image entropy difference into a per-token reward.

\subsection{Chain-of-Thought Compression}
\label{ssec:related_compression}

\begin{figure*}[t]
\centering
\includegraphics[width=0.97\textwidth]{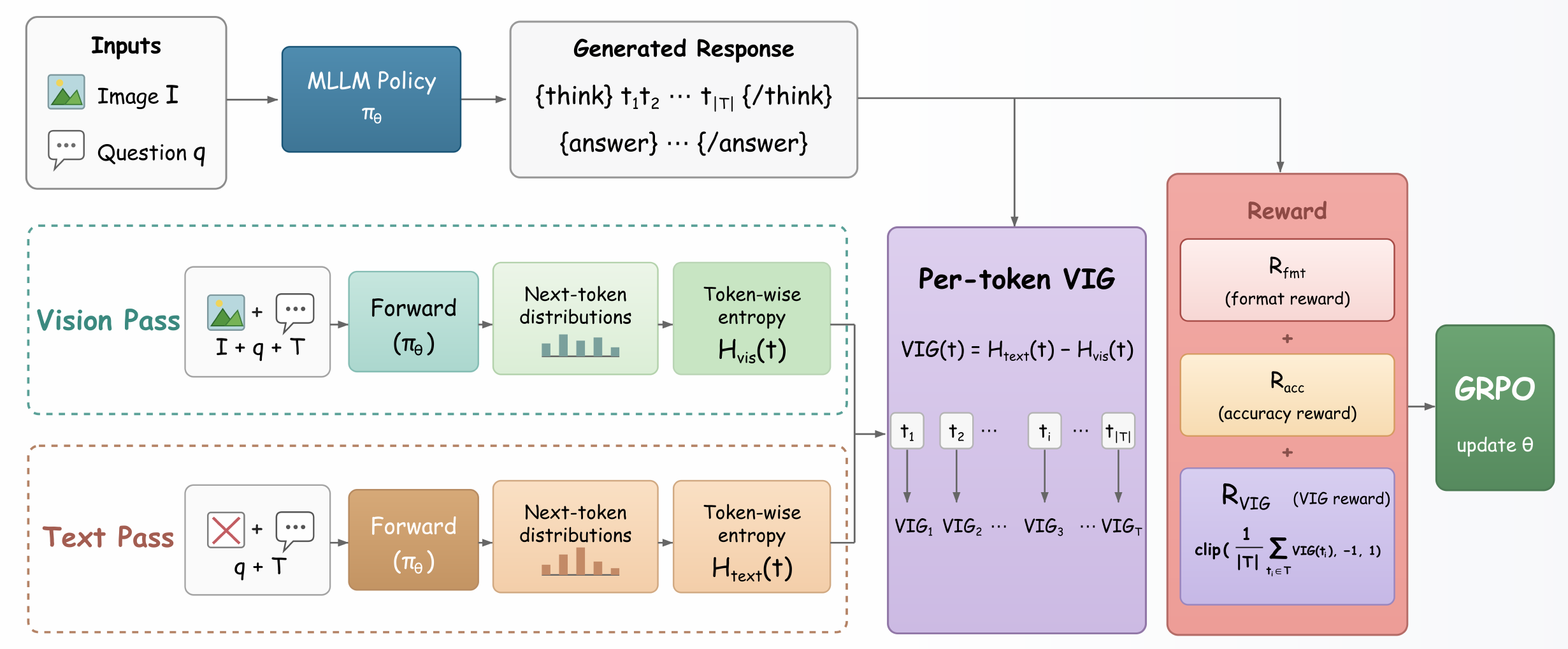}
\caption{\textbf{VIG framework overview.}
Two forward passes over the generated reasoning chain---one with the image
and one with visual tokens removed---yield a token-level entropy difference,
which is aggregated into $R_{\mathrm{VIG}}$ and used as a GRPO reward.}
\label{fig:overview}
\end{figure*}

Existing CoT compression methods fall into three broad families distinguished by
where the compression signal originates.
\emph{Prompt-based} methods control CoT length through input instructions without
modifying model parameters: TALE~\citep{han2025token} estimates a per-question token
budget and embeds it in the system prompt, and Kimi k1.5~\citep{kimi_k1_5_2025}
toggles between long and short reasoning via a budget flag---training-free but
compressing by truncation rather than by selection.
\emph{Training-based} methods fold the compression signal into a supervised or RL
objective.
At the token level, TokenSkip~\citep{tokenskip_2025} ranks tokens by semantic
importance for controllable skipping, while CoT-Valve~\citep{cotvalve_2025}
identifies a parameter-space direction along which chain length elastically varies;
at the step level, StepEntropy~\citep{li2026making} and
Prune-on-Logic~\citep{pruneonlogic_2025} segment the chain and prune low-utility
steps before SFT distillation.
RL-based variants---L1~\citep{l1_2025}, ThinkPrune~\citep{hou2025thinkprune}, and
small-scale preference optimization~\citep{lcpo_2025}---reward shorter outputs
through length-target rewards, iteratively tightened token limits, or
preference-based penalties.
\emph{Latent and architectural} alternatives bypass tokenized CoT entirely:
Coconut~\citep{coconut2024} treats hidden states as continuous
thoughts, LightThinker~\citep{lightthinker_2025} compresses thought steps into
gist tokens via specialized attention masks, the Markovian
Thinker~\citep{markovianthinker_2025} chunks reasoning with explicit carryover
state, and R1-Compress~\citep{r1compress_2025} performs chunk-level inference-time
compression---typically at the cost of architectural changes or the
interpretability of explicit reasoning.

All these approaches share a blind spot for the multimodal setting: their
compression signal is purely textual---token count, text-only entropy, or
text-internal importance---and none measures how much each reasoning token
actually depends on the image.
Information-theoretic objectives are well established in representation
learning~\citep{mine2018,infonce2018} and have been applied through the
information-bottleneck lens to attribute and prune redundant visual
features~\citep{tishby2000information,alemi2017deep,schulz2020restricting},
but always at the level of the visual encoder, not the reasoning chain that
consumes it.
VIG transfers this IB intuition from ``compressing the encoder'' to
``compressing the chain of thought'' by using visual--textual conditional
mutual information as a per-token, online RL reward.

\section{Method}
\label{sec:method}

Our framework, summarized in Figure~\ref{fig:overview}, scores every reasoning
token by its VIG (Eq.~\eqref{eq:vig_def}), aggregates these per-token scores
into a sequence-level reward, and combines it with standard format and
accuracy rewards inside GRPO.
The reward is computed online from two forward passes of the same
policy---one with the image and one with visual tokens removed---so no
reference chains, external annotators, or auxiliary models are required.
Crucially, VIG never rewards shorter outputs directly: it pushes the policy
toward chains in which every token is anchored to the image, and the chain
shortening observed in our experiments emerges as a \emph{byproduct} of this
density-aware optimization.
The remainder of this section defines VIG (\S\ref{ssec:vig_def}) and details
the reward design and two-pass computation (\S\ref{ssec:reward}); we adopt
GRPO~\citep{grpo2024} as the policy-optimization framework and refer the
reader to Appendix~\ref{app:prelim} for the standard background.

\subsection{Visual Information Gain}
\label{ssec:vig_def}

\begin{table*}[t]
\centering
\scriptsize
\setlength{\tabcolsep}{3pt}
\caption{Main results on six benchmarks.
\textbf{Bold} and \underline{underline} denote the best and second-best results
within each model block; tied values share the same marking.}
\label{tab:main}
\resizebox{\textwidth}{!}{%
\begin{tabular}{@{}l|cc|cc|cc|cc|cc|cc|ccc@{}}
\toprule
\multirow{2}{*}{\textbf{Method}}
  & \multicolumn{2}{c|}{\textbf{WeMath}}
  & \multicolumn{2}{c|}{\textbf{MMMU}}
  & \multicolumn{2}{c|}{\textbf{MathVision}}
  & \multicolumn{2}{c|}{\textbf{DynaMath}}
  & \multicolumn{2}{c|}{\textbf{MMK12}}
  & \multicolumn{2}{c|}{\textbf{Geo3K}}
  & \multicolumn{3}{c}{\textbf{Summary}} \\
\cmidrule(lr){2-3}\cmidrule(lr){4-5}\cmidrule(lr){6-7}
\cmidrule(lr){8-9}\cmidrule(lr){10-11}\cmidrule(lr){12-13}\cmidrule(lr){14-16}
  & ACC & TLen & ACC & TLen & ACC & TLen
  & ACC & TLen & ACC & TLen & ACC & TLen
  & AvgACC & AvgTLen & AvgEff \\
\midrule
Qwen3-VL-2B-Thinking
  & 51.72 & 3120 & 36.78 & 2855 & 15.39 & 3928
  & 51.60 & 2252 & 33.25 & 3683 & 33.78 & 3087
  & 37.09 & 3154.2 & 12.71 \\
GRPO
  & 57.36 & 2850 & \textbf{45.00} & 2614 & 17.17 & 3825
  & 54.27 & 2046 & 36.70 & 3592 & \underline{38.60} & 2929
  & 41.52 & 2975.9 & 15.29 \\
L1
  & \textbf{61.67} & \underline{2241} & \underline{43.33} & 2561 & 17.14 & \underline{3668}
  & 53.65 & 1926 & 34.30 & 3429 & 35.94 & \underline{2670}
  & 41.01 & 2749.0 & 16.74 \\
TALE
  & 53.39 & 3033 & 40.00 & 2667 & 17.24 & 3872
  & 54.01 & 2096 & 35.20 & 3595 & 36.11 & 3040
  & 39.32 & 3050.5 & 14.08 \\
ThinkPrune
  & 60.80 & 2438 & 42.56 & \textbf{2372} & \underline{17.53} & 3708
  & \underline{55.53} & \underline{1863} & \textbf{39.70} & \underline{3413} & 37.44 & 2679
  & \underline{42.26} & \underline{2745.5} & \underline{17.17} \\
\textbf{VIG}\,(Ours)
  & \underline{60.98} & \textbf{2159} & \underline{43.33} & \underline{2395} & \textbf{18.72} & \textbf{3639}
  & \textbf{56.55} & \textbf{1728} & \underline{37.20} & \textbf{3317} & \textbf{40.10} & \textbf{2550}
  & \textbf{42.81} & \textbf{2631.3} & \textbf{18.52} \\
\midrule
Qwen3-VL-4B-Thinking
  & 66.84 & 2650 & 53.00 & 2432 & 24.38 & 3657
  & 65.33 & 2012 & 52.05 & 3230 & 54.08 & 2635
  & 52.61 & 2769.6 & 20.46 \\
GRPO
  & 73.79 & 2095 & \underline{57.78} & 2142 & 30.39 & 3378
  & 70.30 & 1714 & 58.00 & 2949 & 61.56 & 2336
  & 58.64 & 2435.5 & 26.37 \\
L1
  & \underline{75.06} & \underline{1876} & 56.78 & 2157 & \textbf{31.71} & 3305
  & 70.70 & 1636 & \underline{60.35} & \underline{2738} & \underline{61.73} & \underline{2114}
  & \underline{59.39} & \underline{2304.4} & \underline{28.40} \\
TALE
  & 71.90 & 2152 & 57.22 & \underline{2002} & 29.21 & 3389
  & 69.40 & \underline{1569} & 54.15 & 3027 & 55.41 & 2506
  & 56.22 & 2440.7 & 25.81 \\
ThinkPrune
  & 74.94 & 1944 & \textbf{58.89} & 2082 & \underline{31.51} & \underline{3276}
  & \underline{71.10} & 1614 & 59.90 & 2817 & 58.07 & 2230
  & 59.07 & 2327.2 & 27.97 \\
\textbf{VIG}\,(Ours)
  & \textbf{79.31} & \textbf{850} & \textbf{58.89} & \textbf{1416} & 30.95 & \textbf{2415}
  & \textbf{73.59} & \textbf{772} & \textbf{65.90} & \textbf{1795} & \textbf{64.23} & \textbf{1310}
  & \textbf{62.15} & \textbf{1426.5} & \textbf{54.79} \\
\midrule
Qwen3-VL-8B-Thinking
  & 73.56 & 1796 & 60.89 & 1955 & 28.85 & 3332
  & 69.04 & 1464 & 59.85 & 2868 & 59.07 & 2075
  & 58.54 & 2248.5 & 29.54 \\
GRPO
  & 79.20 & 1465 & 63.11 & 1779 & 30.63 & 3099
  & \underline{73.87} & 1319 & 66.60 & 2369 & \underline{63.89} & 1821
  & \underline{62.88} & 1975.4 & 36.44 \\
L1
  & 78.51 & 1436 & \underline{63.67} & 1795 & 30.20 & 3022
  & 72.40 & 1339 & \underline{66.75} & 2360 & 62.90 & 1759
  & 62.41 & 1951.8 & 36.38 \\
TALE
  & 76.78 & \underline{1123} & 59.67 & \underline{1557} & \underline{32.50} & \underline{3004}
  & 70.00 & \underline{1013} & 63.40 & \underline{2163} & 59.57 & \textbf{1460}
  & 60.32 & \underline{1720.2} & \underline{42.78} \\
ThinkPrune
  & \underline{79.43} & 1367 & 62.56 & 1690 & \textbf{32.57} & 3038
  & 72.10 & 1263 & 65.40 & 2366 & 63.23 & 1804
  & 62.55 & 1921.4 & 37.60 \\
\textbf{VIG}\,(Ours)
  & \textbf{81.26} & \textbf{947} & \textbf{64.67} & \textbf{1364} & 32.20 & \textbf{2735}
  & \textbf{75.31} & \textbf{892} & \textbf{70.70} & \textbf{1728} & \textbf{65.39} & \underline{1480}
  & \textbf{64.92} & \textbf{1524.3} & \textbf{52.42} \\
\bottomrule
\end{tabular}}
\end{table*}

With this optimization background in place, we next define the central quantity
used to score multimodal reasoning tokens.
Specifically, we ask how much the image reduces the model's uncertainty about a
given token in the reasoning chain.
Consider a single token $t_\tau$ generated at position $\tau$, conditioned on
the question $q$ and the preceding context $\text{ctx}_{<t}=t_{<\tau}$.
By the chain rule of conditional entropy, the predictive uncertainty of $t_\tau$
admits the orthogonal decomposition
\begin{equation}
\begin{aligned}
  H(t_\tau \mid q,\, \text{ctx}_{<t})
  =\;& \underbrace{H(t_\tau \mid q,\, I,\, \text{ctx}_{<t})}_{\hvis(t_\tau)} \\
  +\;& \underbrace{I(t_\tau;\, I \mid q,\, \text{ctx}_{<t})}_{\vig(t_\tau)},
\end{aligned}
\label{eq:decomp}
\end{equation}
where the first term $\hvis(t)$ is the residual uncertainty that remains
\emph{after} observing the image, and the second term is the additional
uncertainty \emph{removed by} the image, i.e., the conditional mutual information
between the token and the visual input.
We refer to the latter as the token's \emph{visual information gain}.

\begin{definition}[Visual Information Gain]
\label{def:vig}
For a token $t_\tau$ at position $\tau$ in the reasoning chain $T$, the
visual information gain is defined as
\begin{equation}
  \vig(t_\tau) \;=\; \htxt(t_\tau) \;-\; \hvis(t_\tau),
  \label{eq:vig_def}
\end{equation}
where $\hvis(t_\tau) = H(t_\tau \mid q, I, t_{<\tau})$ and
$\htxt(t_\tau) = H(t_\tau \mid q, \emptyset, t_{<\tau})$ are the Shannon
entropies of the policy's next-token distribution \emph{with} and
\emph{without} the visual input.
\end{definition}

By the data processing inequality~\citep{cover2006elements},
$\hvis(t_\tau) \le \htxt(t_\tau)$ and $\vig(t_\tau) \ge 0$ almost surely
(Appendix~\ref{app:proof});
Eq.~\eqref{eq:decomp} further identifies $\vig(t_\tau)$ with the per-token
conditional mutual information $I(t_\tau;\, I \mid q, t_{<\tau})$.
Intuitively, a \emph{high-VIG} token is confident with the image but uncertain
without it---the signature of a visually grounded observation---whereas a
\emph{near-zero VIG} token is derivable from the textual context alone.
VIG thus gives a principled per-token measure of how much each reasoning
token depends on the visual evidence.

\subsection{Reward Design}
\label{ssec:reward}

Having defined VIG at the token level, we now show how it is converted into a
sequence-level reward that can be used during policy optimization.
Our reward design proceeds in three stages: aggregating token-level VIG over the
\texttt{<think>} block, computing the resulting score online with a two-pass
forward procedure, and combining it with format and accuracy rewards.

\subsubsection{Token-Level Aggregation}
\label{sssec:aggregation}
We aggregate VIG over all $|T|$ tokens in the \texttt{<think>} block by the
clipped token-level mean
\begin{equation}
  \rvig \;=\; \operatorname{clip}\!\left(
    \frac{1}{|T|}\sum_{t \in T}\vig(t),\; -1,\; 1
  \right).
  \label{eq:rvig_full}
\end{equation}
This choice is principled: by the chain rule of mutual information,
$I(T; I \mid q)=\sum_{t \in T}\vig(t)$, so the unclipped mean is a
length-normalized estimate of the total visual information carried by the
reasoning chain.
In practice, this simple aggregation already provides an effective and stable
reward signal for RL training.

\begin{table*}[t]
\centering
\scriptsize
\setlength{\tabcolsep}{3pt}
\caption{Performance comparison on R1-Onevision-Bench.
\textbf{Bold} and \underline{underline} denote the best and second-best results.}
\label{tab:ood}
\resizebox{\textwidth}{!}{%
\begin{tabular}{@{}l|cc|cc|cc|cc|cc|ccc@{}}
\toprule
\multirow{2}{*}{\textbf{Method}}
  & \multicolumn{2}{c|}{\textbf{Physics} (278)}
  & \multicolumn{2}{c|}{\textbf{Mathematics} (327)}
  & \multicolumn{2}{c|}{\textbf{Biology} (134)}
  & \multicolumn{2}{c|}{\textbf{Chemistry} (105)}
  & \multicolumn{2}{c|}{\textbf{Deduction} (98)}
  & \multicolumn{3}{c}{\textbf{Overall}} \\
\cmidrule(lr){2-3}\cmidrule(lr){4-5}\cmidrule(lr){6-7}
\cmidrule(lr){8-9}\cmidrule(lr){10-11}\cmidrule(lr){12-14}
  & ACC & TLen & ACC & TLen & ACC & TLen
  & ACC & TLen & ACC & TLen
  & AvgACC & AvgTLen & AvgEff \\
\midrule
Qwen3-VL-8B-Thinking
  & 53.96 & 2793 & 45.26 & 2894 & 56.72 & 2121
  & 60.95 & 2315 & 17.35 & 3953
  & 48.30 & 2799.8 & 18.48 \\
GRPO
  & \underline{61.87} & 2444 & 47.09 & 2775 & \textbf{62.69} & 1845
  & 64.76 & 2125 & \underline{24.49} & 3836
  & \underline{53.29} & 2583.1 & 22.62 \\
L1
  & 61.51 & 2338 & 48.32 & 2708 & 59.70 & 1844
  & \textbf{68.57} & 2101 & 21.43 & 3676
  & 53.29 & 2508.9 & 23.00 \\
TALE
  & 56.83 & \underline{2149} & \underline{48.62} & \underline{2565} & \underline{61.94} & \underline{1621}
  & \textbf{68.57} & \underline{1647} & 23.47 & \textbf{3444}
  & 52.55 & \underline{2297.1} & \underline{26.41} \\
ThinkPrune
  & 59.35 & 2416 & 46.48 & 2808 & 58.21 & 1671
  & 63.81 & 1993 & \textbf{26.53} & 3797
  & 51.80 & 2542.8 & 22.99 \\
\textbf{VIG}\,(Ours)
  & \textbf{65.47} & \textbf{1974} & \textbf{54.13} & \textbf{2473} & 60.45 & \textbf{1446}
  & \textbf{68.57} & \textbf{1619} & 23.47 & \underline{3488}
  & \textbf{56.79} & \textbf{2190.0} & \textbf{29.19} \\
\bottomrule
\end{tabular}}
\end{table*}

\subsubsection{Two-Pass Forward Computation}
\label{sssec:two_pass}
$\rvig$ is computed \emph{online} during the GRPO rollout via two forward
passes of the same policy $\pi_\theta$: a \emph{vision pass} on the full
multimodal input $(I,q,s_T)$ yields $\hvis(t)$, and a \emph{text pass} that
removes the visual tokens (the \texttt{<vision>}$\cdots$\texttt{</vision>}
spans) from the same input yields $\htxt(t)$.
The token-level difference, restricted to the \texttt{<think>} block via a
binary mask, is averaged and clipped to $[-1,+1]$.
Both passes share the same weights---no auxiliary parameters or gradients are
introduced---making VIG a drop-in scalar reward.
In practice, the second forward operates on a shorter, vision-stripped input
and does not require gradients, so per-step training time grows by only
$\sim$17\% over plain GRPO (Appendix~\ref{app:cost}).
The full procedure is given as Algorithm~\ref{alg:vig} in
Appendix~\ref{app:algo}.

\subsubsection{Full Reward Function}
\label{sssec:full_reward}

The final reward returned to GRPO combines three signals: format compliance,
answer correctness, and sequence-level visual information gain:
\begin{equation}
\small
\begin{aligned}
  R(o) \,=\;& w_f\,\rfmt(o) + w_a\,\racc(o,a^*) \\
            &+ w_v\,\rvig(o).
\end{aligned}
\label{eq:reward}
\end{equation}
We use $w_f{=}0.1$, $w_a{=}1.0$, and $w_v{=}1.0$ by default; a sensitivity
analysis on $w_v$ is reported in Section~\ref{ssec:ablation}.
Below, we describe the three terms in turn.

\noindent\textbf{Format reward ($\rfmt$).}
This term encourages the model to preserve the desired
\texttt{<think>}/\texttt{<answer>} structure:
\begin{equation}
\small
\rfmt(o)=
\begin{cases}
0 & \text{if both tags are present,}\\
-0.5 & \text{if only \texttt{<answer>} is present,}\\
-1 & \text{otherwise.}
\end{cases}
\label{eq:r_fmt}
\end{equation}

\noindent\textbf{Accuracy reward ($\racc$).}
This term reflects whether the extracted final answer $A(o)$ matches the ground
truth $a^*$:
\begin{equation}
\small
\racc(o,a^*)=
\begin{cases}
1 & \text{if } A(o)=a^*,\\
0 & \text{otherwise.}
\end{cases}
\label{eq:r_acc}
\end{equation}

\noindent\textbf{VIG reward ($\rvig$).}
This term is the clipped mean token-level visual information gain over the
\texttt{<think>} block, as defined in Eq.~\eqref{eq:rvig_full} and computed by
Algorithm~\ref{alg:vig}.
Larger values indicate that a greater fraction of the reasoning chain is
supported by visual evidence rather than by text-only continuation.

Taken together, these three terms guide the model toward responses that are
well-formed, correct, and visually grounded.
In particular, $\rvig$ shapes the reasoning style by rewarding chains whose
tokens remain visually informative on average.
As a result, the policy learns to retain tokens that directly read or relate
visual evidence, while suppressing meta-cognitive openers, repeated problem
restatements, and other visually disengaged filler.
This is how VIG produces shorter and denser reasoning chains without introducing
an explicit length penalty.

\section{Experiments}
\label{sec:exp}

\subsection{Experimental Setup}
\label{ssec:exp_setup}

\paragraph{Baselines.}
We evaluate VIG on Qwen3-VL-2B-Thinking, Qwen3-VL-4B-Thinking, and
Qwen3-VL-8B-Thinking~\citep{bai2025qwen3}.
We compare against five baselines: \textbf{Base}, the original thinking model
without RL fine-tuning; \textbf{GRPO}, trained with only format and accuracy
rewards; \textbf{L1}~\citep{l1_2025}, which uses a length-target RL reward;
\textbf{TALE}~\citep{han2025token}, a dynamic budget prompting method; and
\textbf{ThinkPrune}~\citep{hou2025thinkprune}, which applies RL under an
iteratively tightened token limit.

\paragraph{Evaluation benchmarks.}
We evaluate on seven widely-used multimodal reasoning benchmarks:
WeMath~\citep{wemath_2024}, MMMU~\citep{mmmu_2023},
MathVision~\citep{mathvision_2024}, DynaMath~\citep{dynamath_2024},
MMK12~\citep{mmeureka_2025}, Geo3K~\citep{geo3k_2021}, and
R1-Onevision-Bench~\citep{r1onevision_2025};
the last one provides subject-level evaluation across Physics, Mathematics,
Biology, Chemistry, and Deduction, and is reported separately to highlight
per-subject performance.

\paragraph{Metrics.}
We report accuracy (ACC $\uparrow$), average reasoning-chain length in tokens
(TLen $\downarrow$), and efficiency
$\mathrm{Eff}=\mathrm{ACC}/\mathrm{TLen}\times1000$ ($\uparrow$), which measures
accuracy per kilotoken.
AvgACC and AvgTLen are averaged over the six main benchmarks, while AvgEff is
the mean of per-benchmark efficiency scores.
All numbers we report are from a \emph{single} deterministic (greedy,
temperature 0) decoding pass per checkpoint rather than a mean over sampled
runs, so they carry no decoding variance; where we compare two methods and the
margin matters, we report paired-bootstrap confidence intervals over items
instead (\S\ref{ssec:main_ood}, Appendix~\ref{app:controls}).

\paragraph{Training setup.}
All VIG models are trained directly with GRPO, without an SFT warm-up.
The training set contains 2,948 samples from MMStar~\citep{mmstar_2024}, MathVista~\citep{mathvista_2023}, and LogicVista~\citep{logicvista_2024}.
Unless otherwise specified, VIG uses the no-image comparison pass and $w_v=1.0$.
Detailed dataset statistics, hyperparameters, and implementation settings are
provided in Appendix~\ref{app:exp_details}.

\subsection{Main Results}
\label{ssec:main_ood}

\begin{table}[t]
\centering
\small
\setlength{\tabcolsep}{5pt}
\caption{Evaluation results on non-mathematical visual tasks with
Qwen3-VL-4B-Thinking.
All models use the same checkpoints as the 4B block of Table~\ref{tab:main}.
\textbf{Bold} and \underline{underline} denote the best and second-best
results.}
\label{tab:nonmath}
\begin{tabular}{@{}l|cc|cc@{}}
\toprule
\multirow{2}{*}{\textbf{Method}}
  & \multicolumn{2}{c|}{\textbf{RealWorldQA}  }
  & \multicolumn{2}{c}{\textbf{OCRBench}  } \\
\cmidrule(lr){2-3}\cmidrule(lr){4-5}
  & ACC & TLen & ACC & TLen \\
\midrule
Qwen3-VL-4B-Thinking            & 67.84 & 784  & \underline{77.70} & 587 \\
GRPO            & \underline{68.10} & 635  & 77.30 & 483 \\
L1              & 67.19 & 1116 & 76.00 & 779 \\
ThinkPrune      & \underline{68.10} & \underline{453} & 77.50 & \underline{367} \\
\textbf{VIG}\,(Ours)
                & \textbf{68.63} & \textbf{374} & \textbf{78.30} & \textbf{201} \\
\bottomrule
\end{tabular}
\end{table}

We compare VIG against the base model and four representative compression
baselines---plain GRPO, L1, TALE, and ThinkPrune---across three
Qwen3-VL-Thinking model sizes and six multimodal reasoning benchmarks
(Table~\ref{tab:main}).
Two patterns emerge.
First, length-budget baselines do shorten reasoning chains relative to the
base model, but only marginally improve, and sometimes degrade, average
accuracy: on the 8B backbone, L1 trims AvgTLen from 2249 to 1952 while
AvgACC moves only from 58.54 to 62.41, and TALE actually drops accuracy
to 60.32.
Second, VIG behaves qualitatively differently---rather than trading
accuracy for length, it improves both simultaneously.
On Qwen3-VL-8B-Thinking, VIG attains AvgACC\,=\,64.92\% and
AvgEff\,=\,52.42, surpassing the strongest compression baseline
ThinkPrune~\citep{hou2025thinkprune} by $+2.37\%$ in accuracy and $+39.4\%$
in efficiency, and shortening AvgTLen by an additional 397 tokens.
Compared with plain GRPO~\citep{grpo2024}, which leaves visually-disengaged
elaboration unchecked, VIG cuts the average chain from 1975 to 1524 tokens
($-23\%$) while \emph{raising} accuracy by $+2.04\%$, indicating that the
gains come from removing tokens that did not help answer the question
rather than from stricter length optimization.
The same pattern holds on 4B, where VIG ranks first on AvgACC, AvgTLen, and
AvgEff. VIG also ranks first on all three metrics at 2B, but we state that
accuracy claim more narrowly: the margin over the strongest baseline is only
$+0.55$ points and is not significant under a unified step-400 protocol
($+0.74$), because 53\% of 2B generations hit the 4,096-token limit (vs.\
31.5\% at 8B) with a high repetition rate, so this block is confounded by
runaway-length behaviour (Appendix~\ref{app:controls}).
We therefore read 2B as consistently best efficiency with accuracy on par with
the strongest baseline, and treat 4B/8B as carrying the accuracy claim.

We additionally evaluate VIG on R1-Onevision-Bench~\citep{r1onevision_2025},
which provides subject-level evaluation across Physics, Mathematics, Biology,
Chemistry, and Deduction (Table~\ref{tab:ood}).
On Qwen3-VL-8B-Thinking, VIG attains the best overall accuracy
(56.79) and the best AvgEff (29.19), ranks first on
Physics and Mathematics, and ties for first on Chemistry while using fewer
tokens than every baseline on four of the five subjects.
For instance, VIG outperforms the strongest baseline by $+3.60\%$ on Physics
at $80.8\%$ of its tokens, and by $+5.51\%$ on Mathematics at $\sim96\%$ of
its chain length.
TALE compresses aggressively but trails VIG by $4.24\%$ on overall accuracy,
confirming that the joint accuracy--efficiency improvement is driven by the
visual-density signal rather than by length budgeting.

\paragraph{Beyond mathematics.}

\begin{table}[t]
\centering
\small
\setlength{\tabcolsep}{3.5pt}
\caption{Ablation results on Qwen3-VL-8B-Thinking.
All values are averaged over the six benchmarks of Table~\ref{tab:main}.}
\label{tab:abl}
\begin{tabular}{@{}llccc@{}}
\toprule
\textbf{Factor} & \textbf{Variant} & AvgACC & AvgTLen & AvgEff \\
\midrule
Reward & w/o VIG & 62.88 & 1975.4 & 36.44 \\
       & \textbf{w/ VIG} & \textbf{64.92} & \textbf{1524.3} & \textbf{52.42} \\
\midrule
Vision pass & masked image & 62.98 & 1885.0 & 38.70 \\
            & \textbf{no image} & \textbf{64.92} & \textbf{1524.3} & \textbf{52.42} \\
\midrule
$w_v$ & 0.5 & 63.06 & 1676.1 & 45.04 \\
      & \textbf{1.0} & \textbf{64.92} & \textbf{1524.3} & \textbf{52.42} \\
      & 2.0 & 64.30 & 1546.0 & 50.76 \\
\bottomrule
\end{tabular}
\end{table}

Our benchmark suite is already broader than pure mathematics---MMMU spans 30
subjects across six disciplines, and R1-Onevision covers physics, chemistry, and biology---yet
perception-centric tasks such as visual question answering, scene
understanding, and document understanding remain under-represented.
To test whether the benefit of visual-density-aware optimization carries over
to such tasks, we evaluate the same 4B checkpoints on
RealWorldQA~\citep{realworldqa_2024}, which targets real-world scene
understanding, and OCRBench~\citep{ocrbench_2024}, which targets text and
document recognition, under an identical decoding protocol
(Table~\ref{tab:nonmath}).
The efficiency effect transfers cleanly: VIG produces the shortest chains on
both benchmarks, reducing output tokens by 52\% and 66\% relative to the base
model, while attaining the highest accuracy of all methods.
We read the accuracy result conservatively, since the gaps among the stronger
methods fall within noise (paired-bootstrap intervals in
Appendix~\ref{app:controls}); the claim these results support is that VIG
preserves accuracy at a substantially lower token cost outside mathematical
reasoning, rather than that it improves accuracy there.

\subsection{Ablation Studies}
\label{ssec:ablation}

\begin{figure}[t]
\centering
\includegraphics[width=0.98\columnwidth]{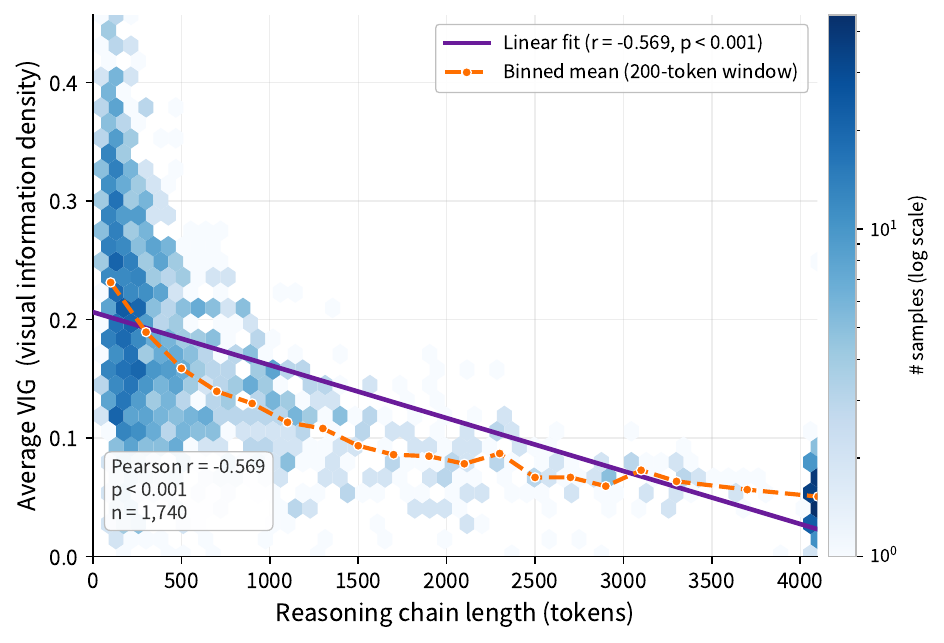}
\caption{Average per-token VIG decreases as reasoning chains lengthen
($r=-0.569$, $p<0.001$): long chains are dominated by visually-disengaged
tokens, exactly the redundancy VIG suppresses.}
\label{fig:vig_length}
\end{figure}

We ablate the three design choices of VIG on Qwen3-VL-8B-Thinking,
averaging across the six benchmarks of Table~\ref{tab:main}
(Table~\ref{tab:abl}).
\textit{(i) The VIG reward.}
Removing the VIG term from the reward reduces the method to plain GRPO,
which collapses AvgEff from 52.42 to 36.44 ($-30\%$) and lowers AvgACC by
2.04 points; conversely, adding the VIG term shortens chains by 451
tokens \emph{while} improving accuracy.
This is the central observation of the ablation: increasing visual
information density and shortening reasoning are not in tension---the
former drives the latter.
\textit{(ii) The text-only contrast pass.}
Replacing the text-only forward pass with a 50\% masked-image pass
($H_{\mathrm{vis}}^{\mathrm{mask}}\!-\!H_{\mathrm{vis}}$) weakens the
information-theoretic contrast, costing 1.94 points of accuracy and 13.72
points of efficiency, which validates the conditional-mutual-information
formulation: removing visual context entirely produces a sharper signal
than partially masking it.
\textit{(iii) Reward weight $w_v$.}
A sweep over $w_v\!\in\!\{0.5,1.0,2.0\}$ traces an inverted-U on AvgEff:
too small under-rewards visual grounding, while too large over-emphasizes
the visual-density signal at the expense of the accuracy reward---accuracy
drops by 0.62 points without producing any meaningful further chain
shortening, since TLen stays around 1.5k once visually-disengaged tokens
have been suppressed.
$w_v\!=\!1.0$ is the joint optimum and is used throughout the rest of the
paper.

\begin{figure}[tb]
\centering
\includegraphics[width=0.98\columnwidth]{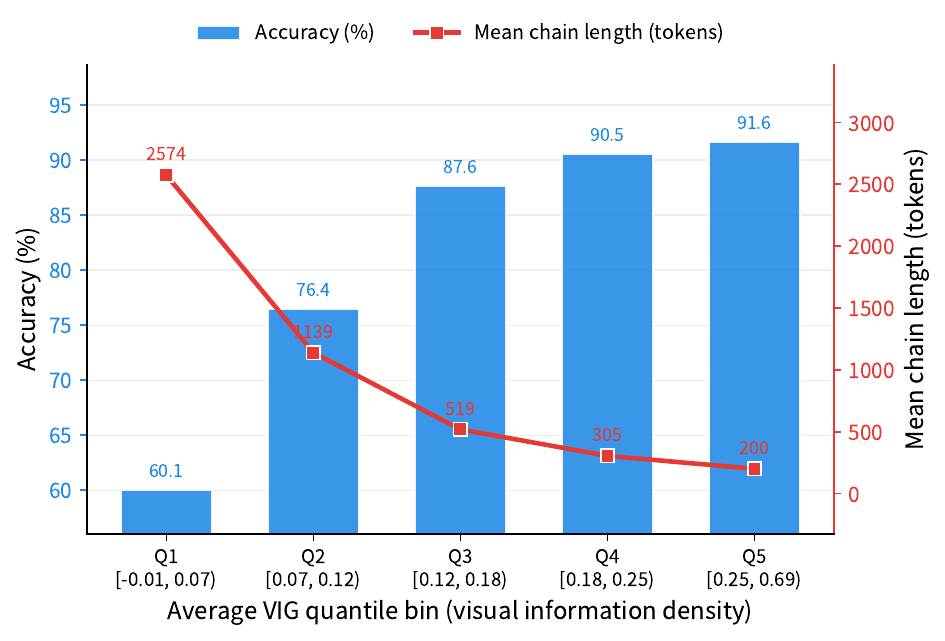}
\caption{Accuracy and mean chain length stratified by per-sample average
VIG (WeMath, $n=1{,}740$): higher visual density co-occurs with both higher
accuracy and shorter chains.
This association is descriptive; it is partly mediated by chain length and
item difficulty (\S\ref{ssec:visualization}).}
\label{fig:acc_vig}
\end{figure}

\subsection{Visualization Analysis}
\label{ssec:visualization}

To understand \emph{why} VIG simultaneously improves accuracy and shortens
reasoning, we examine the relationship between visual information density
and chain behavior on a held-out subset of WeMath generations.

\subsubsection{Visual Density vs. Chain Length}
We first investigate how per-token visual information density evolves as
reasoning chains grow (Figure~\ref{fig:vig_length}, $\sim$1.7k VIG-trained
generations).
A clear negative trend emerges (Pearson $r=-0.569$, $p<0.001$): as chains
grow longer, the average per-token visual information gain decreases
monotonically.
In other words, the additional tokens that lengthen a chain are
disproportionately tokens whose distributions can be predicted from
linguistic context alone---meta-cognitive openers, problem restatements,
self-verification, generic algebra---rather than tokens grounded in the
image.
This is precisely the failure mode that uniform length penalties cannot
target, and it is the reason VIG's reward, which directly upweights
visually-grounded tokens, induces shorter chains as a \emph{byproduct}
of optimizing for visual density rather than as an explicit length
constraint.

\subsubsection{Visual Density vs. Accuracy}
We next stratify the same generations into VIG quintiles and examine how
accuracy and chain length co-vary with visual density
(Figure~\ref{fig:acc_vig}).
Both quantities move monotonically with visual density: bins with higher
average VIG attain substantially higher accuracy
(Q1\,$\to$\,Q5: 60.1\%\,$\to$\,91.7\%) and use markedly fewer tokens
(Q1\,$\to$\,Q5: 2574\,$\to$\,200; intermediate quintiles are
76.4\%/1139, 87.6\%/519, and 90.5\%/305).

We read this analysis as \emph{descriptive rather than causal}: a
partial-correlation check shows the sample-level ACC--VIG association is
largely mediated by chain length and item difficulty (partial $r\approx0$ for
correctness after controlling both, while the VIG--length relation survives
difficulty control at partial $r=-0.63$).
The causal evidence that raising visual density \emph{produces} this behaviour
is therefore carried by the controlled ablations of \S\ref{ssec:ablation} and
Appendix~\ref{app:controls}, not by this correlation.
With that qualification, the two analyses indicate that VIG acts not as a
length budget but as a mechanism that discards visually-irrelevant
elaboration, yielding the chains documented in
Tables~\ref{tab:main}--\ref{tab:abl}.

\section{Conclusion}
\label{sec:conclusion}

We introduced \textbf{VIG} (Visual Information Gain), an information-theoretic
reward for multimodal chain-of-thought compression that measures how much each
reasoning token depends on the image.
Unlike existing compression methods that optimize for shorter outputs directly,
VIG trains the model to concentrate its reasoning on visually informative
content, thereby improving efficiency by increasing visual density rather than by
imposing a length budget.
The reward is fully online, requires no reference chains or external annotators,
and integrates naturally with GRPO through a two-pass forward computation.
Across six main multimodal reasoning benchmarks and three Qwen3-VL-Thinking
model sizes, plus an additional R1-Onevision-Bench evaluation on 8B, VIG
consistently improves the accuracy--efficiency trade-off and ranks first
overall and first or tied-first on three of five R1-Onevision subject
categories.
Overall, our results suggest that efficient multimodal CoT generation requires
not merely thinking less, but making each token rely more on visual evidence.

\section*{Limitations}
\label{sec:limitations}

VIG has several important limitations.
Computing the reward requires an additional text-only forward pass during RL
training, which raises optimization cost relative to standard GRPO, and it
depends on a clean separation between visual and textual tokens, so its
implementation is somewhat model-specific.

Most importantly, VIG measures \emph{visual dependence}, which is neither
\emph{visual correctness} nor \emph{counterfactual necessity}.
A score is high whenever the image substantially shifts the policy's next-token
distribution, so a confidently hallucinated reading can score as high as
genuinely grounded perception; the accuracy reward counteracts this only at the
trajectory level, and combining VIG with explicit grounding signals is a
natural extension.
Nor does high VIG imply indispensability: as Appendix~\ref{app:controls} shows,
deleting low-VIG sentences from an already-compressed chain hurts more than
deleting high-VIG ones, because what survives at low VIG is largely
load-bearing computation.
VIG is therefore suited to shaping the output distribution during training, but
not to post-hoc pruning.

Finally, our study covers static image reasoning with Qwen3-VL-Thinking models,
and we regard cross-architecture generality as \emph{unresolved}: a rigorous
cross-family study would require matched thinking-mode backbones and per-family
re-tuning of the RL stack, and a single under-tuned transfer run would be weak
evidence either way.
We release our vision-span abstraction, which lets other families be plugged in
without changing the reward, to facilitate such verification.

\bibliography{references}

\appendix

\section{Preliminaries}
\label{app:prelim}

\subsection{Multimodal CoT}
\label{app:prelim_mmcot}
Given a visual question $(I, q)$ composed of an image $I$ and a textual query $q$,
a modern Multimodal Large Reasoning Model produces a structured response of the
form
\[
  \scalebox{0.85}{$\displaystyle
  \underbrace{\langle\texttt{think}\rangle\; t_1 \cdots t_{|T|}\; \langle\texttt{/think}\rangle}_{\text{reasoning chain }T}
  \;\underbrace{\langle\texttt{answer}\rangle\; a \;\langle\texttt{/answer}\rangle}_{\text{final answer}}
  $}
\]
where the \texttt{<think>} block contains an ordered sequence of reasoning
tokens $T=(t_1,\ldots,t_{|T|})$, and $a$ is the final answer.
VIG operates exclusively on the reasoning chain $T$, leaving the format and answer
portions unaffected.

\subsection{GRPO}
\label{app:prelim_grpo}
To optimize the policy, we adopt Group Relative Policy Optimization
(GRPO)~\citep{grpo2024}, which replaces a learned value baseline with
intra-group reward normalization.
For each prompt $(I,q)$, the old policy $\pi_{\theta_\text{old}}$ samples a group
of $G$ responses $\{o_1,\ldots,o_G\}$ with scalar rewards $\{r_1,\ldots,r_G\}$.
The group-relative advantage assigned to the $i$-th sample is
\begin{equation}
  \hat{A}_{i} \,=\,
  \frac{r_i - \operatorname{mean}(\{r_j\}_{j=1}^{G})}
       {\operatorname{std}(\{r_j\}_{j=1}^{G})}.
  \label{eq:grpo_adv}
\end{equation}
Let
$\rho_{i,t}(\theta)=\pi_\theta(o_{i,t}\mid q,I,o_{i,<t})/\pi_{\theta_\text{old}}(o_{i,t}\mid q,I,o_{i,<t})$
be the importance ratio at token $t$.
GRPO then maximizes the clipped surrogate objective
\begin{equation}
\small
\mathcal{J}_{\mathrm{GRPO}}(\theta)
= \mathbb{E}_{(q,I),\,\{o_i\}}
\left[ \frac{1}{G}\sum_{i=1}^{G}\frac{1}{|o_i|}
\sum_{t=1}^{|o_i|} \mathcal{L}_{i,t} \right],
\label{eq:grpo}
\end{equation}
where the token-level term is
\begin{equation}
\small
\begin{aligned}
\mathcal{L}_{i,t}
=\;& \min\Big(\rho_{i,t}\hat{A}_i,
\operatorname{clip}(\rho_{i,t},1-\varepsilon,1+\varepsilon)\hat{A}_i\Big) \\
&- \beta\, D_{\mathrm{KL}}\big(\pi_\theta\,\Vert\,\pi_{\mathrm{ref}}\big).
\end{aligned}
\label{eq:grpo_loss}
\end{equation}
Here $(q,I)\sim\mathcal{D}$ and $\{o_i\}\sim\pi_{\theta_\text{old}}$, while
$\varepsilon$ is the PPO-style clipping range and $\beta$ controls the KL
regularization strength.
Under this framework, all task-specific supervision enters through the scalar
reward $r_i$, which is exactly where our VIG-based reward is introduced.

\section{Two-Pass VIG Computation}
\label{app:algo}

Algorithm~\ref{alg:vig} details how $\rvig$ is computed online during each
GRPO rollout.
For every sampled response $o=(s_T, a)$ produced by $\pi_\theta$, the
controller assembles two parallel inputs from the same prompt and chain:
\texttt{vis\_inputs}, the standard multimodal input that includes the visual
token spans \texttt{<vision>}$\cdots$\texttt{</vision>}; and
\texttt{txt\_inputs}, identical except that all visual spans have been
stripped out so the model conditions purely on textual context.
Both inputs are batched and passed through $\pi_\theta$ in a single forward
call, so the cost of computing $\rvig$ is two forward passes per rollout
sample---no extra weights, no backward pass, and no auxiliary models.

From the resulting logits we compute per-token entropies
$\hvis(t)=-\sum_v p_v^{\mathrm{vis}}\log p_v^{\mathrm{vis}}$ and
$\htxt(t)=-\sum_v p_v^{\mathrm{txt}}\log p_v^{\mathrm{txt}}$, then take their
position-wise difference.
A binary mask restricts the difference to positions inside the
\texttt{<think>} block; positions corresponding to format tags, the
\texttt{<answer>} segment, and any padding are excluded so that the reward
reflects only the reasoning chain.
The masked sequence is averaged across positions and clipped to $[-1,+1]$ to
absorb the rare floating-point negative values discussed in
Appendix~\ref{app:proof}.
The resulting scalar enters the composite reward of Eq.~\eqref{eq:reward}
without any gradient flowing through the two forward passes, so VIG behaves
as a pure reward signal rather than a differentiable auxiliary loss.

\begin{algorithm}[t]
\caption{Computing $\rvig$ for a sampled response}
\label{alg:vig}
\small
\begin{algorithmic}[1]
\Input model $\pi_\theta$, processor, image $I$, question $q$, think text $s_T$
\Output scalar $\rvig \in [-1, +1]$
\State \textbf{[Vision pass]} $\gets$ build \texttt{vis\_inputs} from $(I, q, s_T)$
\State \hspace{1.8em} Run forward: $\text{logits}_\text{vis} \gets \pi_\theta(\text{vis\_inputs})$
\State \hspace{1.8em} $\hvis(t) \gets -\sum_v p_v \log p_v$ \quad (per token)
\State \textbf{[Text pass]} Remove all \texttt{<vision>}$\cdots$\texttt{</vision>} tokens
\State \hspace{1.8em} Run forward: $\text{logits}_\text{txt} \gets \pi_\theta(\text{txt\_inputs})$
\State \hspace{1.8em} $\htxt(t) \gets -\sum_v p_v \log p_v$ \quad (per token)
\State \textbf{[Aggregate]}
  $\text{mask} \gets$ positions in $\langle\texttt{think}\rangle$ block
\State \hspace{1.8em} $\vig\_\text{seq} \gets (\htxt - \hvis)[\text{mask}]$
\State \hspace{1.8em} $\rvig \gets \operatorname{clip}\!\left(\operatorname{mean}(\vig\_\text{seq}),\;-1,\;+1\right)$
\end{algorithmic}
\end{algorithm}

\section{Proof of VIG Non-Negativity}
\label{app:proof}

\paragraph{Population statement.}
Let $p$ denote the true joint distribution over
$(t_\tau, q, I, \text{ctx}_{<t})$.
The conditional mutual information satisfies
$I(t_\tau;\,I \mid q,\,\text{ctx}_{<t}) \geq 0$ with equality iff $t_\tau$
is conditionally independent of $I$ given $(q,\,\text{ctx}_{<t})$
\citep{cover2006elements}.
Equivalently, conditioning on additional information cannot increase entropy:
\begin{equation}
  H(t_\tau \mid q,\, I,\, \text{ctx}_{<t})
  \;\leq\; H(t_\tau \mid q,\, \text{ctx}_{<t}),
  \label{eq:cond_ent}
\end{equation}
so $\hvis(t_\tau) \leq \htxt(t_\tau)$ and $\vig(t_\tau) \geq 0$ at the
population level.

\paragraph{Estimator and sources of negative values.}
In practice $\hvis$ and $\htxt$ are not population entropies but
\emph{plug-in estimates} obtained from the policy $\pi_\theta$:
$\hat{H}_{\text{vis}}(t_\tau) = -\sum_{v} \pi_\theta(v\mid q, I, \text{ctx}_{<t})
\log \pi_\theta(v\mid q, I, \text{ctx}_{<t})$ and analogously for
$\hat{H}_{\text{txt}}$ with the visual tokens stripped from the input.
Because $\pi_\theta$ is not the true conditional---and, crucially, because the
text-only pass evaluates the same multimodal-trained network on an
out-of-distribution input---the empirical entropies need not satisfy the
population inequality above.
We identify three sources of occasional negative empirical VIG:

\begin{enumerate}[leftmargin=*,nosep]
  \item \textit{Policy mis-specification.} $\pi_\theta$ is a learned model,
    not the data-generating distribution; the data-processing inequality is
    a statement about $p$, not about $\pi_\theta$, so noisy estimates can
    flip sign on individual tokens.
  \item \textit{Out-of-distribution text-only pass.} The shared weights
    encode multimodal training; removing the visual tokens places the
    network outside its training distribution, occasionally inducing
    higher-entropy or attention-leakage behaviour
    (\emph{phantom} attention to absent vision positions).
  \item \textit{Numerical noise.} Computing entropy in float16/bf16 with
    \texttt{F.log\_softmax} introduces $\mathcal{O}(10^{-3})$ rounding error;
    in well-grounded tokens this can manifest as small negative VIG.
\end{enumerate}

Empirically, negative values are rare and concentrated near zero
(less than $\sim$1\% of think-block tokens, $|\vig|<0.05$ for the vast
majority of the affected tokens).
We therefore clip $\rvig$ to $[-1,+1]$ and average over the
\texttt{<think>} block, which renders the reward stable while preserving the
information-theoretic interpretation of Eq.~\eqref{eq:decomp} at the
population level.

\section{Data Details}
\label{app:exp_details}

To comprehensively evaluate multimodal CoT compression, we use seven reasoning
benchmarks in total: six benchmarks for the main comparison plus one
additional benchmark with subject-level evaluation.
For RL training, we construct the training mixture from three datasets: MMStar,
MathVista, and LogicVista.
Below we summarize the role and characteristics of each dataset used in this
paper.

\paragraph{Evaluation benchmarks.}
\begin{itemize}[nosep,leftmargin=1.2em]
  \item \textbf{WeMath}~\citep{wemath_2024}: a multimodal mathematics benchmark designed to study how large multimodal models solve visual math problems rather than only measuring end-to-end performance. We evaluate on 1,740 examples.
  \item \textbf{MMMU}~\citep{mmmu_2023}: a college-level multimodal benchmark spanning diverse disciplines and requiring broad knowledge together with strong visual understanding. We evaluate on 900 examples.
  \item \textbf{MathVision}~\citep{mathvision_2024}: a benchmark for multimodal mathematical reasoning over diagrams, formulas, charts, and geometry figures. We evaluate on 3,040 examples.
  \item \textbf{DynaMath}~\citep{dynamath_2024}: a visual mathematical reasoning benchmark designed to test robustness under dynamic perturbations and distribution shifts. We evaluate on 5,010 examples.
  \item \textbf{MMK12}~\citep{mmeureka_2025}: a multi-subject K--12 multimodal mathematics benchmark with human-verified answers, introduced alongside MM-Eureka. We evaluate on the 2,000-example test split released by \citet{papo_2025}.
  \item \textbf{Geo3K}~\citep{geo3k_2021}: a geometry benchmark centered on diagram understanding and symbolic reasoning. We evaluate on 601 examples.
  \item \textbf{R1-Onevision-Bench}~\citep{r1onevision_2025}: a comprehensive multimodal reasoning benchmark covering mathematics, physics, chemistry, biology, and deduction. We use all 942 examples for subject-level evaluation.
\end{itemize}

\paragraph{RL training data.}
\begin{itemize}[nosep,leftmargin=1.2em]
  \item \textbf{MMStar}~\citep{mmstar_2024}: a challenging multimodal benchmark emphasizing fine-grained perception, spatial reasoning, and image--text alignment. We sample from its validation split for RL training.
  \item \textbf{MathVista}~\citep{mathvista_2023}: a diverse multimodal mathematics benchmark covering charts, geometry, symbolic reasoning, and visual word problems. We use its testmini split for RL training.
  \item \textbf{LogicVista}~\citep{logicvista_2024}: a benchmark for multimodal logical reasoning in visual contexts, containing problems that require multi-step deduction grounded in images. We use its test split for RL training.
\end{itemize}
In total, the RL training mixture contains 2,948 examples.

\section{Implementation Details}
\label{app:impl}

\paragraph{Baseline implementation.}
\textbf{Base} directly uses the original Qwen3-VL-Thinking checkpoints without
additional post-training.
\textbf{GRPO}~\citep{grpo2024} uses the same training pipeline as VIG but removes
the VIG reward, retaining only the format and accuracy terms.
\textbf{L1}~\citep{l1_2025} follows the same GRPO framework with a
length-oriented reward, \textbf{TALE}~\citep{han2025token} is implemented as a
training-free prompting baseline with a dynamic token budget, and
\textbf{ThinkPrune}~\citep{hou2025thinkprune} uses a pruning-oriented RL signal
based on reasoning-length control.
For fair comparison, all RL-based baselines use the same training data mixture,
model backbones, and evaluation protocol as VIG.

\paragraph{VIG implementation.}
For each sampled response, VIG computes token-level visual information gain by
running the same policy twice: once with the image and once with the visual
tokens removed.
In Qwen3-VL, visual tokens are enclosed by
\texttt{<|vision\_start|>} and \texttt{<|vision\_end|>} markers; the text-only
pass removes the full span between these markers.
The resulting entropy difference is restricted to the \texttt{<think>} block,
averaged over tokens, and clipped to obtain $R_{\mathrm{VIG}}$.
Unless otherwise specified, all main experiments use token-level aggregation,
the no-image comparison pass, and reward weight $w_v=1.0$.

\paragraph{Training configuration.}
All models are trained with full-parameter GRPO using group size $G=8$,
temperature $1.0$, KL coefficient $0.04$, and at most 500 optimization steps.
We use batch size 1 per device, gradient accumulation 4, learning rate $10^{-6}$,
warmup steps 10, gradient checkpointing, and DeepSpeed ZeRO-2.
The optimal checkpoint for VIG is selected at step 400 in the main experiments;
baselines use step 500 for Qwen3-VL-2B-Thinking and Qwen3-VL-8B-Thinking, and
step 400 for Qwen3-VL-4B-Thinking.

\paragraph{Numerical details.}
Entropy is computed in nats using \texttt{F.log\_softmax} for numerical stability.
In practice, VIG values are clipped to $[-1,1]$ before being used as rewards.

\section{Signal-Integrity Controls, Deletion Probe, and Scope}
\label{app:controls}

This appendix reports the controls that establish \emph{why} the VIG term
produces the gains of \S\ref{ssec:ablation}, a probe that delimits what the
signal measures, and the analyses that bound the scope of our claims.

\subsection{Protocols}
\label{app:protocols}

\paragraph{Confidence-only arm.}
This arm shares the pipeline, data, seed, and step count of VIG, with the
reward's visual term replaced by $-\hvis$ computed over the same
\texttt{<think>} mask.
It therefore rewards low predictive entropy \emph{with} the image and is by
construction insensitive to the visual contrast, which isolates whether the
gain could come from entropy or brevity shaping alone.

\paragraph{Stratification.}
For every evaluation item we run the untrained base model with the image
removed under otherwise identical decoding settings.
Items it still answers correctly form the \emph{text-solvable} stratum and the
remainder the \emph{vision-required} stratum; VIG and GRPO are then re-scored
within each stratum, with confidence intervals from a 10k paired bootstrap over
items.

\paragraph{Sentence typing.}
Sentences are labelled \emph{visual} when they contain image-referential
expressions, \emph{algebra} when they contain arithmetic or symbolic
manipulation, \emph{meta} for metacognitive openers and self-checks, and
\emph{other} otherwise, with \emph{visual} taking precedence when a sentence
qualifies for both.
Over the 70{,}754 sentences of VIG-8B WeMath outputs, mean VIG by type is
0.153 (visual, $n{=}10{,}821$), 0.106 (meta, $6{,}780$), 0.088
(other, $32{,}524$), and 0.065 (algebra, $20{,}629$).

\paragraph{Deletion probe.}
Sentences of a completed chain are ranked by VIG, the stated fraction is
removed, and the surviving text is supplied as a forced prefix from which the
answer is re-decoded greedily.
We report two conditions, with the image still in context and with the image
also removed at re-decoding.
The VIG-chain probe covers rates $\{10,20,30,40,50\}\%$ and the GRPO-chain
probe $\{25,50\}\%$.

\subsection{Interpreting the Deletion Asymmetry}
\label{app:asymmetry}

Table~\ref{tab:deletion} shows that on VIG-trained chains, deleting the
lowest-VIG sentences is the most damaging condition, the opposite of what a
token-importance reading of VIG would predict.
Two mechanisms combine to produce this.

The first is content type.
Low-VIG sentences are enriched in algebra and derivation, which is visually
ungrounded yet computation-bearing, so removing it breaks the arithmetic.
A high-VIG sentence, by contrast, is typically a readout of a value that
remains recoverable at re-decoding time because the image is still in context;
consistent with this, high-VIG deletion costs only $1.67$ points even when the
image is \emph{also} removed.

The second is chain state.
The asymmetry is specific to chains produced by a policy already trained for
visual density.
On chains from a GRPO-trained policy the ordering reverses, so the reward does
track answer-critical visual content when that content has not yet been
compressed away.
One confound is worth stating explicitly: low-VIG sentences are somewhat longer
on average, so deleting a fixed fraction of \emph{sentences} removes more
tokens in that condition, although the strict ordering holds at every rate.

\subsection{Scope of the Comparison}
\label{app:scope}

\paragraph{Trajectory-level reweighting (VPPO).}
VPPO~\citep{vppo_2025} is the closest contemporaneous method, reweighting
trajectory advantages by an overall visual-dependency measure rather than
scoring individual tokens.
Under our protocol, VIG reaches 71.37 AvgACC at 1{,}297 tokens versus 63.39 at
1{,}100 tokens for the released VPPO-8B checkpoint: VPPO produces slightly
shorter chains and therefore a higher efficiency ratio, while VIG retains
substantially higher accuracy.
We treat this as \emph{indicative only} rather than an apples-to-apples
comparison, because the available checkpoint is built on a different backbone
(Qwen3-VL-8B-Instruct rather than the Thinking variant) and trained on
different data; a fully matched reproduction is left to future work.

\paragraph{Failure cases.}
MathVision, the hardest and longest-chain benchmark in our suite, is the one
benchmark where VIG does not lead accuracy: ThinkPrune (32.57) and TALE
(32.50) edge past VIG (32.20) at 8B, and L1 leads at 4B (31.71 vs.\ 30.95).
The margins lie within noise ($\leq0.8$ points) but are consistent across
scales, while VIG retains by far the shortest chains (2{,}735 vs.\
$\geq$3{,}004 tokens at 8B).
Our reading is that when a problem genuinely requires long exploration, the
density signal has less redundancy to remove and mild compression can trim
useful search.
Two further limits are worth stating.
On 2B, 53\% of generations hit the 4,096-token limit (versus 31.5\% at 8B) with
a high repetition rate, so accuracy differences in that block are dominated by
runaway-length behaviour.
And because $\rvig$ is a chain-level mean, a chain that hinges on one small
visual detail can be under-rewarded relative to one with many mildly visual
sentences, as illustrated in Figure~\ref{fig:appendix_heatmap}(b).
At more generous decoding budgets (8k tokens, no truncation pressure), accuracy
differences among RL recipes narrow while VIG's efficiency advantage persists.

\begin{table}[t]
\centering
\small
\setlength{\tabcolsep}{5pt}
\caption{Signal-integrity controls on Qwen3-VL-8B-Thinking.
All arms share identical data, seed, and step count.
\textbf{(a)} Reward ablation, 5-benchmark macro-average.
\textbf{(b)} VIG vs.\ GRPO split by whether the base model solves the item
without the image, 6-benchmark macro-average.}
\label{tab:controls}
\begin{tabular}{@{}lcc@{}}
\multicolumn{3}{@{}l}{\textbf{(a) Reward ablation}} \\
\toprule
\textbf{Reward} & AvgACC & AvgTLen \\
\midrule
GRPO (no visual term)      & 69.33 & 1751 \\
Confidence-only ($-\hvis$) & 68.21 & 1441 \\
\textbf{VIG (ours)}        & \textbf{71.37} & \textbf{1297} \\
\bottomrule
\end{tabular}

\vspace{6pt}
\begin{tabular}{@{}lccc@{}}
\multicolumn{4}{@{}l}{\textbf{(b) Stratified accuracy}} \\
\toprule
\textbf{Stratum} & VIG & GRPO & $\Delta$ \\
\midrule
Text-solvable    & 83.18 & 81.69 & $+1.49$ \\
Vision-required  & 56.01 & 53.19 & $\mathbf{+2.82}$ \\
\bottomrule
\end{tabular}
\end{table}

\begin{table}[t]
\centering
\small
\setlength{\tabcolsep}{5pt}
\caption{Quantile-deletion probe (WeMath, $n=1{,}740$).
Sentences are ranked by VIG, a fraction is deleted, and the answer is
re-decoded from the surviving prefix with the image still in context.
\textbf{Bold} marks the most damaging condition per block.}
\label{tab:deletion}
\begin{tabular}{@{}lccccc@{}}
\multicolumn{6}{@{}l}{\textbf{(a) VIG-8B chains}} \\
\toprule
\textbf{Deleted} & 10\% & 20\% & 30\% & 40\% & 50\% \\
\midrule
lowest-VIG  & 80.92 & 79.94 & 78.22 & 75.17 & \textbf{72.36} \\
random      & 81.84 & 80.92 & 80.06 & 79.25 & 77.41 \\
highest-VIG & 82.07 & 81.95 & 81.78 & 81.32 & 80.75 \\
\bottomrule
\end{tabular}

\vspace{6pt}
\begin{tabular}{@{}lcc@{}}
\multicolumn{3}{@{}l}{\textbf{(b) GRPO-8B chains}} \\
\toprule
\textbf{Deleted} & 25\% & 50\% \\
\midrule
lowest-VIG  & 81.21 & 80.00 \\
random      & 80.57 & 79.48 \\
highest-VIG & 80.46 & \textbf{75.92} \\
\bottomrule
\end{tabular}
\end{table}

\begin{figure}[t]
\centering
\scriptsize
\setlength{\tabcolsep}{2pt}
{\tiny sentence VIG:~\tikz\fill[gray!18](0,0)rectangle(0.26,0.15); ${<}0$~\tikz\fill[myblue!8](0,0)rectangle(0.26,0.15); ${\le}0.1$~\tikz\fill[myblue!20](0,0)rectangle(0.26,0.15); ${\le}0.25$~\tikz\fill[myblue!35](0,0)rectangle(0.26,0.15); ${\le}0.5$~\tikz\fill[myblue!55](0,0)rectangle(0.26,0.15); ${>}0.5$}

\vspace{4pt}
\textbf{(a)} \emph{There are three paths from point M to point N, one of which is the shortest. This path is ( ).}\\
{\tiny Options: A. A; B. B; C. C; D. Cannot be determined; E. No correct answer \quad$\vert$\quad gold \textbf{B}, model \textbf{B} }

\vspace{2pt}
\begin{minipage}[t]{0.30\columnwidth}
\vspace{0pt}
\centering
\includegraphics[width=\linewidth]{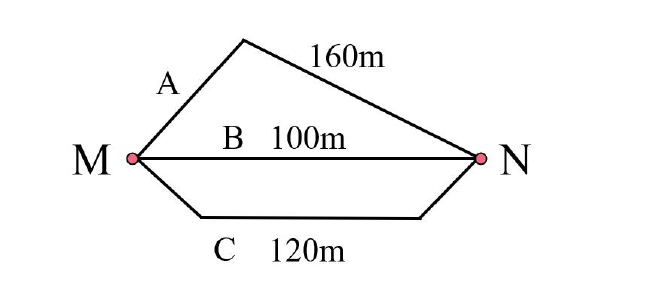}
\end{minipage}\hfill
\begin{minipage}[t]{0.67\columnwidth}
\vspace{0pt}
\begin{tabular}{@{}r@{~}l@{~}p{0.47\columnwidth}@{}}
\toprule
VIG & & Reasoning sentence \\
\midrule
\rowcolor{myblue!20} +0.14 & \tikz[baseline=-0.55ex]\fill[myblue!75!black](0,0)rectangle(0.087cm,0.15cm); & So, let's see. \\
\rowcolor{myblue!8} +0.04 & \tikz[baseline=-0.55ex]\fill[myblue!75!black](0,0)rectangle(0.025cm,0.15cm); & The problem is about finding the shortest path from M to N. \\
\rowcolor{myblue!8} +0.04 & \tikz[baseline=-0.55ex]\fill[myblue!75!black](0,0)rectangle(0.027cm,0.15cm); & There are three paths: A, B, C. \\
\rowcolor{myblue!55} +0.55 & \tikz[baseline=-0.55ex]\fill[myblue!75!black](0,0)rectangle(0.342cm,0.15cm); & Let's check their lengths. \\
\rowcolor{myblue!55} +1.16 & \tikz[baseline=-0.55ex]\fill[myblue!75!black](0,0)rectangle(0.728cm,0.15cm); & Path A: The label says 160m. \\
\rowcolor{myblue!35} +0.44 & \tikz[baseline=-0.55ex]\fill[myblue!75!black](0,0)rectangle(0.277cm,0.15cm); & Path B: 100m. \\
\rowcolor{myblue!35} +0.28 & \tikz[baseline=-0.55ex]\fill[myblue!75!black](0,0)rectangle(0.177cm,0.15cm); & Path C: 120m. \\
\rowcolor{myblue!20} +0.13 & \tikz[baseline=-0.55ex]\fill[myblue!75!black](0,0)rectangle(0.083cm,0.15cm); & So comparing 100, 120, 160, the shortest is B. \\
\rowcolor{myblue!35} +0.45 & \tikz[baseline=-0.55ex]\fill[myblue!75!black](0,0)rectangle(0.282cm,0.15cm); & So the answer should be B. \\
\bottomrule
\end{tabular}
\end{minipage}

\vspace{7pt}
\textbf{(b)} \emph{The perimeter of rectangle ABCD is 20 cm. What is the length of AB in cm?}\\
{\tiny Options: A. 4; B. 3; C. 2; D. 5; E. No correct answer \quad$\vert$\quad gold \textbf{A}, model \textbf{A} }

\vspace{2pt}
\begin{minipage}[t]{0.22\columnwidth}
\vspace{0pt}
\centering
\includegraphics[width=\linewidth]{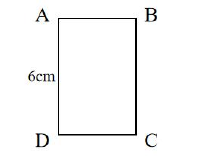}
\end{minipage}\hfill
\begin{minipage}[t]{0.75\columnwidth}
\vspace{0pt}
\begin{tabular}{@{}r@{~}l@{~}p{0.47\columnwidth}@{}}
\toprule
VIG & & Reasoning sentence \\
\midrule
\rowcolor{gray!18} -0.10 & \tikz[baseline=-0.55ex]\fill[gray!60](0,0)rectangle(0.060cm,0.15cm); & So, let's solve this. \\
\rowcolor{myblue!55} +0.73 & \tikz[baseline=-0.55ex]\fill[myblue!75!black](0,0)rectangle(0.454cm,0.15cm); & First, the rectangle has sides AD and BC as 6 cm (since~$\ldots$ \\
\rowcolor{myblue!20} +0.21 & \tikz[baseline=-0.55ex]\fill[myblue!75!black](0,0)rectangle(0.134cm,0.15cm); & The perimeter of a rectangle is 2*(length + width). \\
\rowcolor{myblue!35} +0.29 & \tikz[baseline=-0.55ex]\fill[myblue!75!black](0,0)rectangle(0.179cm,0.15cm); & Let's denote AB as the length, which is the top side, and~$\ldots$ \\
\rowcolor{myblue!8} +0.09 & \tikz[baseline=-0.55ex]\fill[myblue!75!black](0,0)rectangle(0.059cm,0.15cm); & The perimeter is 20 cm. \\
\rowcolor{myblue!20} +0.10 & \tikz[baseline=-0.55ex]\fill[myblue!75!black](0,0)rectangle(0.065cm,0.15cm); & So formula: 2*(AB + AD) = 20. \\
\rowcolor{myblue!8} +0.05 & \tikz[baseline=-0.55ex]\fill[myblue!75!black](0,0)rectangle(0.029cm,0.15cm); & We know AD is 6, so plug that in: 2*(AB + 6) = 20. \\
\rowcolor{myblue!8} +0.01 & \tikz[baseline=-0.55ex]\fill[myblue!75!black](0,0)rectangle(0.020cm,0.15cm); & Divide both sides by 2: AB + 6 = 10. \\
\rowcolor{myblue!20} +0.10 & \tikz[baseline=-0.55ex]\fill[myblue!75!black](0,0)rectangle(0.065cm,0.15cm); & Then AB = 10 - 6 = 4. \\
\rowcolor{myblue!55} +0.98 & \tikz[baseline=-0.55ex]\fill[myblue!75!black](0,0)rectangle(0.613cm,0.15cm); & So the answer should be A. \\
\bottomrule
\end{tabular}
\end{minipage}

\caption{Sentence-level VIG within complete VIG-8B chains, shown beside the
input image.
Sentences that read a quantity off the figure score highest ($+1.16$, $+0.73$),
while symbolic manipulation scores near zero even when answer-bearing
(``Divide both sides by 2'', $+0.01$).}
\label{fig:vig_heatmap}
\end{figure}

\section{Training Cost}
\label{app:cost}

VIG requires a second, text-only forward pass per rollout sample, so its
theoretical worst-case overhead is $2\times$ the cost of plain GRPO.
In practice the gap is substantially smaller because (i)~the text pass
runs on a shorter sequence (the visual-token spans are removed before the
forward), (ii)~only logits are needed---no backward pass and no optimizer
update for the second pass, and (iii)~loading the rollout, computing
advantages, and stepping the optimizer all amortize across the two passes.

We report end-to-end training cost on Qwen3-VL-8B-Thinking, all jobs
trained on a single $8\times$ H20 node for 500 GRPO steps with the same
batch size, rollout length, and dataset.

\begin{table}[h]
\centering
\small
\setlength{\tabcolsep}{4pt}
\caption{Training cost on Qwen3-VL-8B-Thinking (8$\times$H20, 500 GRPO
steps). \emph{Rel.} is the relative wall-clock cost vs.\ plain GRPO.}
\label{tab:cost}
\begin{tabular}{@{}lrrrr@{}}
\toprule
\textbf{Method} & \textbf{s/it} & \textbf{Wall-clock} & \textbf{GPU-h} & \textbf{Rel.} \\
\midrule
GRPO                   & 31.8 & 4h\,25m & 35.3 & $1.00\times$ \\
L1                     & 33.5 & 4h\,39m & 37.2 & $1.05\times$ \\
ThinkPrune             & 32.3 & 4h\,29m & 35.8 & $1.01\times$ \\
\midrule
\textbf{VIG (Ours)}    & 37.2 & 5h\,10m & 41.3 & $1.17\times$ \\
\bottomrule
\end{tabular}
\end{table}

VIG adds only $\sim$17\% to per-step wall-clock relative to plain GRPO, and
remains comparable to other length-based RL baselines such as L1
($1.05\times$) and ThinkPrune ($1.01\times$), because the second forward
operates on a shorter, vision-stripped input and does not require gradient
computation.
We did not observe sample-efficiency degradation: VIG converges within the
same 500-step budget as the other baselines and reaches its final accuracy
without additional rollouts.
We therefore view the cost as a modest, predictable overhead that buys an
information-theoretic reward signal otherwise unavailable through purely
textual compression objectives.

\paragraph{When is the overhead repaid?}
The overhead is a one-time training cost, whereas the token savings recur at
every inference request, so it is useful to state the break-even point
explicitly.
Training VIG costs $41.3-35.3=6.0$ additional GPU-hours.
At inference under our unified 4,096-token protocol, VIG emits 1{,}297 output
tokens per query versus 1{,}751 for plain GRPO, a saving of ${\sim}454$ tokens
per request.
Taking a deliberately conservative aggregate decode throughput of 1{,}000
output tokens per second per GPU for batched serving of an 8B model on this
hardware, 6.0 GPU-hours corresponds to ${\sim}21.6$M generated tokens, so the
extra training cost is repaid after roughly
$21.6\text{M}/454 \approx 48\text{K}$ inference requests.
Beyond that point VIG is strictly cheaper end-to-end than plain GRPO while
also being more accurate.
The estimate scales linearly with the assumed throughput, so a faster serving
stack shortens the break-even point proportionally; the qualitative conclusion
is that the overhead is amortized within a small fraction of a deployed
model's lifetime.

\section{Supplementary Visualization Examples}
\label{app:vis_examples}

\begin{figure}[t]
\centering
\scriptsize
\setlength{\tabcolsep}{2pt}
{\tiny sentence VIG:~\tikz\fill[gray!18](0,0)rectangle(0.26,0.15); ${<}0$~\tikz\fill[myblue!8](0,0)rectangle(0.26,0.15); ${\le}0.1$~\tikz\fill[myblue!20](0,0)rectangle(0.26,0.15); ${\le}0.25$~\tikz\fill[myblue!35](0,0)rectangle(0.26,0.15); ${\le}0.5$~\tikz\fill[myblue!55](0,0)rectangle(0.26,0.15); ${>}0.5$}

\vspace{4pt}
\textbf{(a) Success.} \emph{As shown in the figure, use a ruler to measure the length of the rectangle's side. What is the length of BC in cm?}\\
{\tiny Options: A. 3; B. 4; C. 5; D. No correct answer \quad$\vert$\quad gold \textbf{A}, model \textbf{A} }

\vspace{2pt}
\begin{minipage}[t]{0.22\columnwidth}
\vspace{0pt}
\centering
\includegraphics[width=\linewidth]{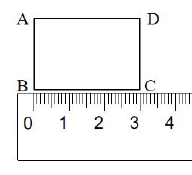}
\end{minipage}\hfill
\begin{minipage}[t]{0.75\columnwidth}
\vspace{0pt}
\begin{tabular}{@{}r@{~}l@{~}p{0.47\columnwidth}@{}}
\toprule
VIG & & Reasoning sentence \\
\midrule
\rowcolor{myblue!8} +0.08 & \tikz[baseline=-0.55ex]\fill[myblue!75!black](0,0)rectangle(0.050cm,0.15cm); & So, let's look at the figure. \\
\rowcolor{myblue!55} +0.58 & \tikz[baseline=-0.55ex]\fill[myblue!75!black](0,0)rectangle(0.360cm,0.15cm); & The rectangle has side BC. \\
\rowcolor{myblue!55} +0.90 & \tikz[baseline=-0.55ex]\fill[myblue!75!black](0,0)rectangle(0.564cm,0.15cm); & The ruler is below it, with markings from 0 to 4. \\
\rowcolor{myblue!35} +0.27 & \tikz[baseline=-0.55ex]\fill[myblue!75!black](0,0)rectangle(0.167cm,0.15cm); & Point B is at the 0 mark, and point C is at the 3 mark. \\
\rowcolor{myblue!8} +0.00 & \tikz[baseline=-0.55ex]\fill[myblue!75!black](0,0)rectangle(0.020cm,0.15cm); & So the length of BC is 3 cm. \\
\rowcolor{myblue!20} +0.12 & \tikz[baseline=-0.55ex]\fill[myblue!75!black](0,0)rectangle(0.076cm,0.15cm); & Let's check the options. \\
\rowcolor{myblue!35} +0.26 & \tikz[baseline=-0.55ex]\fill[myblue!75!black](0,0)rectangle(0.164cm,0.15cm); & Option A is 3. \\
\bottomrule
\end{tabular}
\end{minipage}

\vspace{7pt}
\textbf{(b) Failure.} \emph{As shown in the figure, using a protractor to measure one of the angles of the triangular glass fragment, the measured angle is ()$^\circ$?}\\
{\tiny Options: A. 40; B. 50; C. 60; D. No correct answer \quad$\vert$\quad gold \textbf{A}, model \textbf{B} }

\vspace{2pt}
\begin{minipage}[t]{0.31\columnwidth}
\vspace{0pt}
\centering
\includegraphics[width=\linewidth]{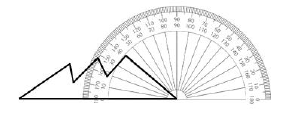}
\end{minipage}\hfill
\begin{minipage}[t]{0.66\columnwidth}
\vspace{0pt}
\begin{tabular}{@{}r@{~}l@{~}p{0.47\columnwidth}@{}}
\toprule
VIG & & Reasoning sentence \\
\midrule
\rowcolor{myblue!20} +0.21 & \tikz[baseline=-0.55ex]\fill[myblue!75!black](0,0)rectangle(0.132cm,0.15cm); & So, let's look at the protractor in the image. \\
\multicolumn{3}{@{}c@{}}{\tiny$\vdots$} \\
\rowcolor{myblue!35} +0.34 & \tikz[baseline=-0.55ex]\fill[myblue!75!black](0,0)rectangle(0.212cm,0.15cm); & First, check the scale. \\
\rowcolor{myblue!35} +0.30 & \tikz[baseline=-0.55ex]\fill[myblue!75!black](0,0)rectangle(0.189cm,0.15cm); & The protractor has two scales, inner and outer. \\
\multicolumn{3}{@{}c@{}}{\tiny$\vdots$} \\
\rowcolor{myblue!20} +0.13 & \tikz[baseline=-0.55ex]\fill[myblue!75!black](0,0)rectangle(0.078cm,0.15cm); & The slanted side of the triangle points to 50$^\circ$ on the~$\ldots$ \\
\rowcolor{myblue!20} +0.14 & \tikz[baseline=-0.55ex]\fill[myblue!75!black](0,0)rectangle(0.088cm,0.15cm); & Wait, let's confirm. \\
\multicolumn{3}{@{}c@{}}{\tiny$\vdots$} \\
\rowcolor{myblue!8} +0.01 & \tikz[baseline=-0.55ex]\fill[myblue!75!black](0,0)rectangle(0.020cm,0.15cm); & The other side of the triangle is at 50$^\circ$. \\
\rowcolor{myblue!35} +0.40 & \tikz[baseline=-0.55ex]\fill[myblue!75!black](0,0)rectangle(0.250cm,0.15cm); & So the measured angle is 50$^\circ$. \\
\rowcolor{myblue!8} +0.00 & \tikz[baseline=-0.55ex]\fill[myblue!75!black](0,0)rectangle(0.020cm,0.15cm); & So the answer is B. \\
\bottomrule
\end{tabular}
\end{minipage}

\caption{Additional cases ($\vdots$: omitted sentences).
(a) Reading the ruler scores $+0.90$, the answer-bearing arithmetic $+0.004$.
(b) A failure case: every sentence is visually dependent, yet the model
misreads the protractor as $50^\circ$ instead of $40^\circ$.
High VIG certifies image-driven generation, not a correct reading.}
\label{fig:appendix_heatmap}
\end{figure}

\begin{figure*}[t]
\centering
\includegraphics[width=0.97\textwidth]{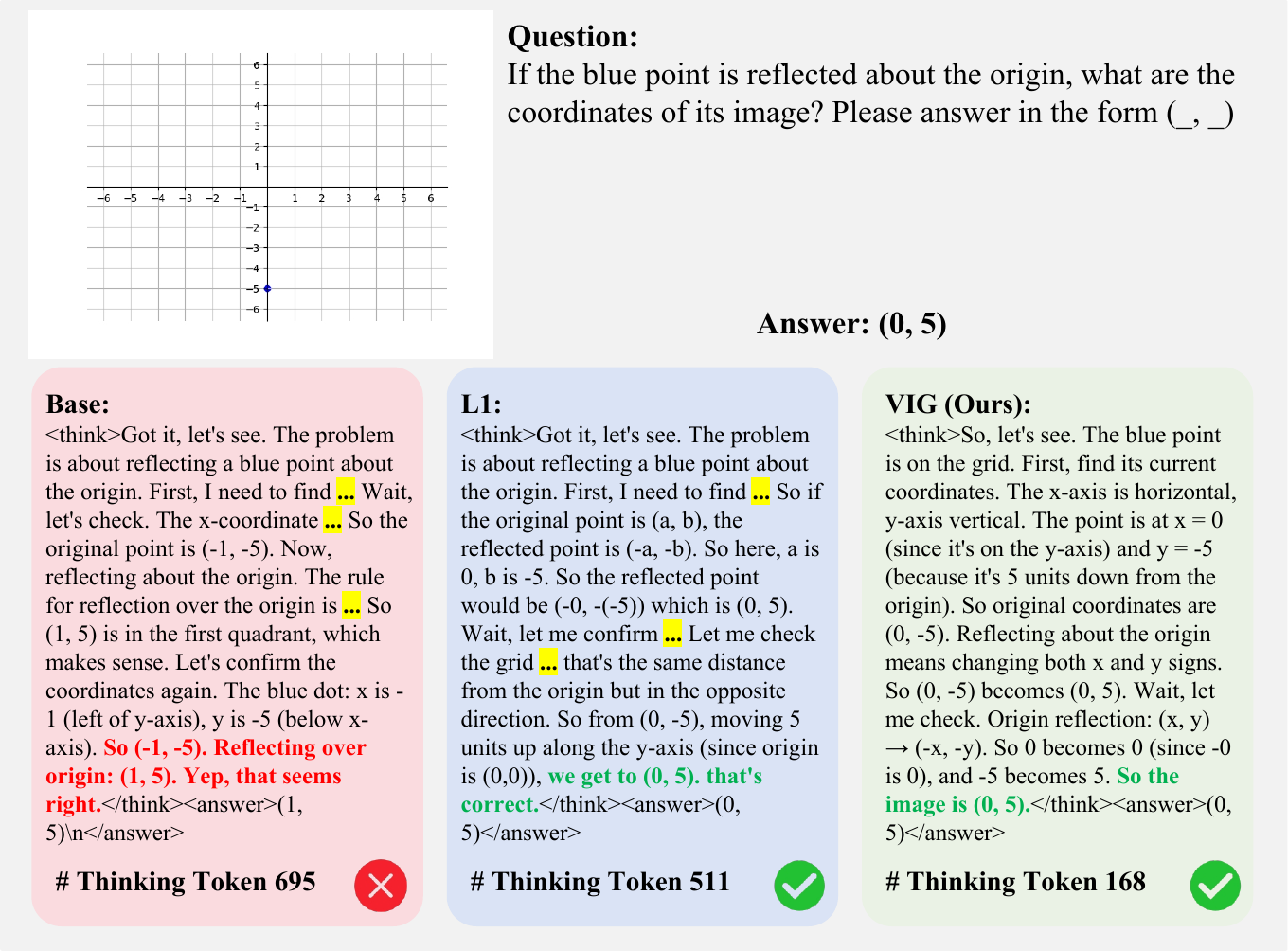}
\caption{An additional qualitative comparison between Base, L1, and VIG.
VIG produces a shorter reasoning chain while preserving the key visual evidence
needed for the final answer.}
\label{fig:appendix_case1}
\end{figure*}

\begin{figure*}[t]
\centering
\includegraphics[width=0.97\textwidth]{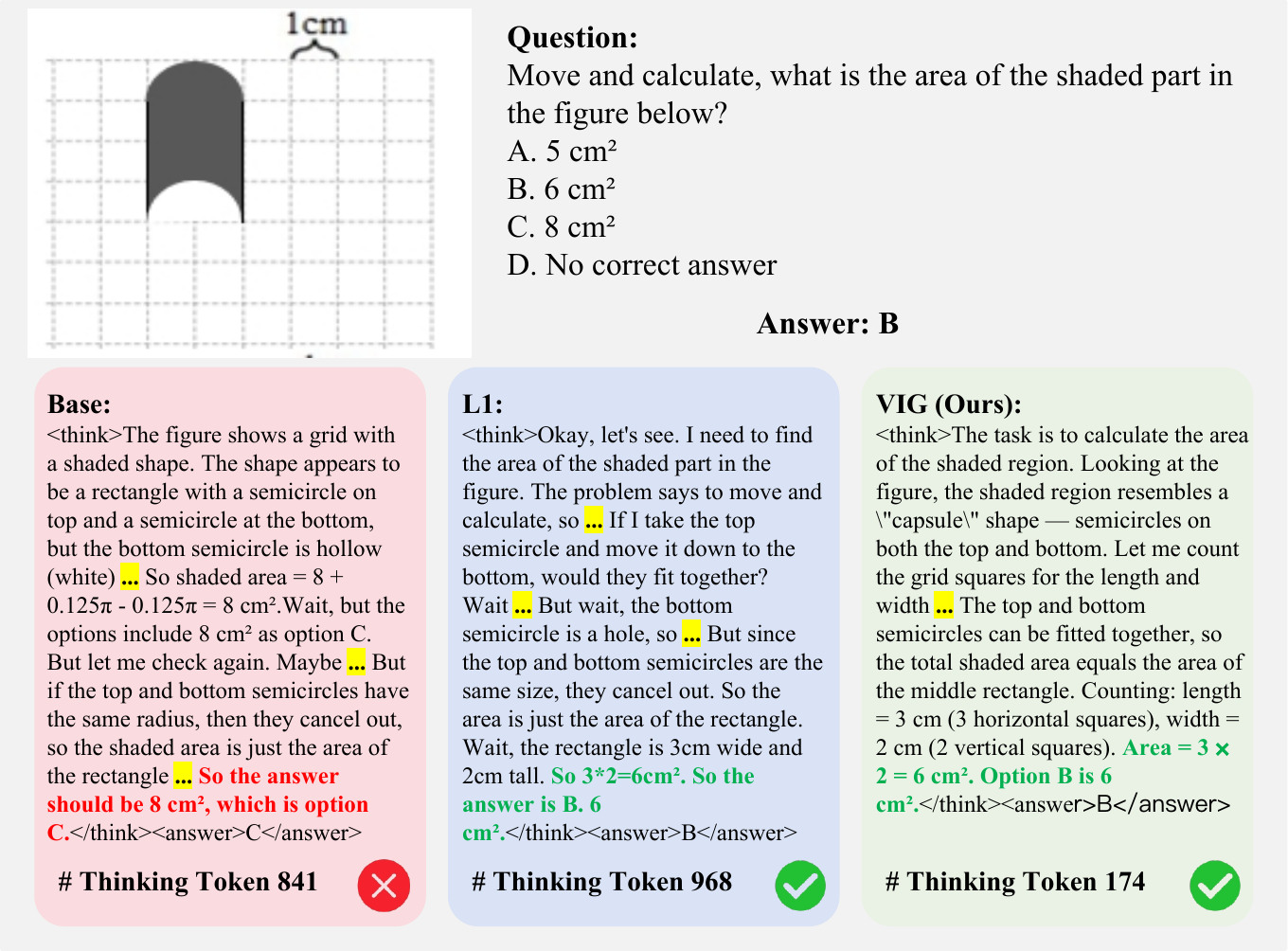}
\caption{A second qualitative example comparing Base, L1, and VIG.
Compared with the baseline methods, VIG better concentrates its reasoning on the
visually relevant information and avoids unnecessary self-reflection.}
\label{fig:appendix_case2}
\end{figure*}

This appendix contains two kinds of qualitative material: additional
sentence-level VIG colourings (Figure~\ref{fig:appendix_heatmap}), and
side-by-side output comparisons against baselines
(Figures~\ref{fig:appendix_case1}--\ref{fig:appendix_case2}).

To complement the motivating example in Figure~\ref{fig:intro}, we also provide
two case studies comparing the base model, L1, and VIG.
These examples illustrate the same qualitative pattern observed throughout the
paper: the base model tends to over-elaborate and drift away from decisive visual
evidence, L1 shortens the chain but does not always improve visual grounding, and
VIG produces shorter reasoning that remains tightly anchored to the image.

\section{Datasets and Licenses}
\label{app:licenses}

All evaluation benchmarks used in this work are publicly released for research
use. WeMath~\citep{wemath_2024}, MathVision~\citep{mathvision_2024},
DynaMath~\citep{dynamath_2024}, Geo3K~\citep{geo3k_2021}, MMMU~\citep{mmmu_2023},
MMK12~\citep{mmeureka_2025,papo_2025}, and R1-Onevision-Bench~\citep{r1onevision_2025}
are distributed under research-permissive licenses (Apache-2.0, MIT, or
CC-BY-{NC-}4.0 variants); our use is restricted to non-commercial academic
evaluation, consistent with their stated intended use.
The two additional perception-centric benchmarks used in
\S\ref{ssec:main_ood} are also publicly available:
RealWorldQA~\citep{realworldqa_2024}, released by xAI alongside the Grok-1.5V
announcement and distributed on Hugging Face under CC-BY-ND-4.0, and
OCRBench~\citep{ocrbench_2024}, released with the MultimodalOCR project.
We use the public splits without modification and report only aggregate
evaluation metrics, consistent with their stated intended use.
The Qwen3-VL-Thinking 2B/4B/8B base models~\citep{bai2025qwen3} are released
under the Qwen license and used here only for research.
None of the benchmarks contain personally identifying information or offensive
content: all items are mathematical, scientific, or geometric reasoning
problems in English.

\section{Use of AI Assistants}
\label{app:ai_use}

We used AI assistants (ChatGPT and Claude) in a strictly limited capacity.
Their use was confined to (i) language polishing of the manuscript---improving
grammar, phrasing, and consistency of terminology---and (ii) minor coding
assistance for non-research utilities such as plotting scripts, LaTeX
formatting, and small data-processing snippets.
All research ideas, the VIG formulation, the algorithmic design, the training
pipeline, the experimental protocol, and the analysis are the authors' own.
No AI assistant was used to generate experimental results, write the
information-theoretic derivations, or produce any scientific claim in this
paper.

\end{document}